\documentclass[review, times, 10pt]{elsarticle}

\usepackage{amssymb}
\usepackage{amsmath,amsfonts}

\usepackage{caption}
\usepackage{algorithmic}
\usepackage{algorithm}
\usepackage{array}
\usepackage{xcolor} 
\usepackage{makecell}
\usepackage[caption=false,font=normalsize,labelfont=sf,textfont=sf]{subfig}

\usepackage[margin=1.5in]{geometry}

\usepackage{lineno}

\journal{Pattern Recognition}

\usepackage{tabularx}
\usepackage{adjustbox}

 \usepackage{threeparttable}

 \usepackage[section]{placeins}

\usepackage{natbib}

\begin{document}

\begin{frontmatter}



\title{Semantic-embedded Similarity Prototype for Scene Recognition}

\author[label1,label2]{Chuanxin Song}
\ead{songchuanxin@mail.sdu.edu.cn}
\author[label1,label2]{Hanbo Wu}
\ead{wuhanbo@sdu.edu.cn}
\author[label1,label2]{Xin Ma\corref{mycorrespondingauthor}}
\ead{maxin@sdu.edu.cn}
\author[label1,label2]{Yibin Li}
\ead{liyb@sdu.edu.cn}

\affiliation[label1]{organization={Center for Robotics, School of Control Science and Engineering, Shandong University},
            country={China}}
\affiliation[label2]{organization={Engineering Research Center of Intelligent Unmanned System, Ministry of Education},
            country={China}}
\cortext[mycorrespondingauthor]{Corresponding author}

\begin{abstract}
Due to the high inter-class similarity caused by the complex composition and the co-existing objects across scenes, numerous studies have explored object semantic knowledge within scenes to improve scene recognition. However, a resulting challenge emerges as object information extraction techniques require heavy computational costs, thereby burdening the network considerably. This limitation often renders object-assisted approaches incompatible with edge devices in practical deployment. In contrast, this paper proposes a semantic knowledge-based similarity prototype, which can help the scene recognition network achieve superior accuracy without increasing the computational cost in practice. It is simple and can be plug-and-played into existing pipelines. More specifically, a statistical strategy is introduced to depict semantic knowledge in scenes as class-level semantic representations. These representations are used to explore correlations between scene classes, ultimately constructing a similarity prototype. Furthermore, we propose to leverage the similarity prototype to support network training from the perspective of Gradient Label Softening and Batch-level Contrastive Loss, respectively. Comprehensive evaluations on multiple benchmarks show that our similarity prototype enhances the performance of existing networks, all while avoiding any additional computational burden in practical deployments. Code and the statistical similarity prototype will be available at https://github.com/ChuanxinSong/SimilarityPrototype
\end{abstract}


\end{frontmatter}


\section{Introduction}
\begin{figure}[htbp]
    \centering
    \includegraphics[width=9cm]{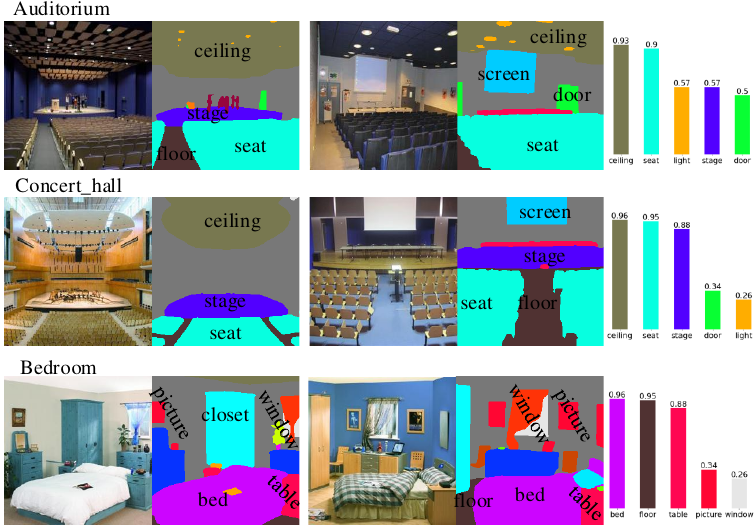}
    \caption{Similarity due to object co-occurrence in scene recognition. The rightmost column represents the probability statistics for the top five occurrences of the object in the scene. Obviously, "Auditorium" and "Concert\_hall" are extremely similar, while "Bedroom" is different from them.}
    \label{fig1}
\end{figure}

Scene recognition is a prominent research area in computer vision, with applications in autonomous driving, robotics, and surveillance \cite{xie2020scene}. Deep Neural Networks (DNNs) \cite{he2016deepr5,howard2019searchingr7,liu2021swinr10,liu2022convnetr12} have achieved significant success in image classification due to their powerful feature extraction capabilities. However, the coexisting objects across scenes pose challenges for feature differentiation. In Fig. \ref{fig1}, for example, similar object distributions result in high similarity between "Auditorium" and "Concert\_hall." This inter-class similarity limits the performance of DNNs in scene recognition \cite{xie2015hybridr13,cheng2018scener19}. To address this issue, many researchers have explored incorporating object semantic knowledge within scenes to enhance scene recognition performance.

In order to use the internal object information to assist in scene recognition, several researchers \cite{sun2018fusingr17,cheng2018scener19,song2019imager20,li2021place} have employed object detection to extract specific object regions from a given scene. Subsequently, they integrate these regional features to identify the scene. However, focusing on objects alone can only cover part of the scenes; the background elements, such as indoor carpets and outdoor sands, also play an important role in recognizing scenes. Recognizing this limitation, alternative strategies \cite{lopez2020semanticr21,zeng2020amorphousr22,song2023srrmr45} have turned to semantic segmentation to extract pixel-level region information within a scene. By modeling the contextual relationships between object regional features, these methods achieve improved results in scene recognition. Despite considerable performance, when applying the trained model in practice, these object-assisted approaches require fine-grained object information extraction from images using object detection or semantic segmentation techniques. These techniques typically require far more computational costs than normal classification networks. This not only places a heavier computational burden on resource-constrained edge devices, but also substantially impedes the speed of scene recognition in real-world applications. Therefore, there is a need to explore more efficient ways of using object information to enhance scene recognition without exacerbating the computational demands.

Exploring the feature context within a scene guided by object information is no longer suitable for the above purposes. Therefore, this paper shifts the focus from the intra-scene object region context to the overall inter-scene correlation. A human perspective inspires our motivation: The high similarity between different classes is a challenge in scene recognition, but not all categories have a high similarity. Take Fig. \ref{fig1} as an example, an "Auditorium" should be more similar to a "Concert\_hall" than to a "Bedroom." Thus, humans learn scene categories by consciously expending more effort on distinguishing those scenes that are more similar. Inspired by this, we argue that if the network is informed of the inter-class similarity during training, it can focus more on distinguishing similar scene classes and thus find appropriate fitting directions faster. Indeed, due to parameter redundancy \cite{hou2024network}, the backbone network possesses substantial potential for achieving improved performance when provided with meticulous training guidance. Recent research on scene essence \cite{qiu2021essence} points out that the similarity between different scenes is often determined by the coexisting objects in them. Consequently, we propose to embed object semantic knowledge into a similarity prototype. This prototype can provide the network with prior knowledge of inter-class similarity during the training process, thereby effectively improving the accuracy of the network.

Specifically, we utilize semantic segmentation techniques and statistical analysis to capture the distribution of object semantic information in training data, subsequently generating class-level semantic representations. These representations are used to quantify the similarity between scene classes, ultimately constructing an inter-class semantic similarity prototype. We propose to use this prototype as prior knowledge to assist model training from two perspectives: label smoothing and contrastive loss. Label smoothing \cite{szegedy2016rethinkingr23} is a regularization method aimed at mitigating overconfidence by replacing hard labels with softer labels (probabilities dispersed across all categories). However, it treats all non-target categories equally, disregarding variations in their relevance to the target category. Contrastive loss \cite{chopra2005learningr29} is a metric learning method that aims to establish an embedding space where similar instances are pulled together while dissimilar instances are pushed apart. Yet, the distance between positive and negative pairs is equal, failing to reflect genuine disparities between similar and dissimilar samples. Fortunately, our similarity prototype offers a solution to these challenges. From the label smoothing perspective, we propose a Gradient Label Softening (GLS) strategy. Concretely, we embed the similarity prototype into softened labels and propose a confidence gradient strategy to gradually guide the model's fitting direction during training. From the contrastive loss perspective, we propose a plug-and-play Batch-level Contrastive Loss (BCL) strategy. Specifically, we use our similarity prototype to measure the inter-class similarity thresholds. We further extend this strategy to intra-class contrastive loss, ensuring the comprehensive use of semantic prior knowledge during training.

Our contributions can be summarized as follows:
\begin{enumerate}
    \item We propose a semantic-embedded similarity prototype to augment the network's training process by providing prior knowledge. The trained network achieves superior recognition performance in real applications without the need for intensive object information extraction, thus effectively reducing computational costs.
    \item We propose two strategies to use our similarity prototype to assist network training: Gradient Label Softening and Batch-level Contrastive Loss. The former embeds our prototype into softened labels and uses a confidence gradient strategy to guide training. The latter uses our prototype to measure the similarity requirements of inter- and intra-class samples in each mini-batch. These two strategies ensure the full use of the semantic knowledge within similarity prototype.
    \item For the first time, our similarity prototype demonstrates that object information could successfully improve scene recognition performance without additional computational burden in practice. Comprehensive evaluations on several publicly available datasets \cite{quattoni2009recognizingr38,xiao2010sunr39,zhou2017placesr40} show that our approach can achieve performance comparable to existing state-of-the-art methods.
\end{enumerate}

\section{Related works}
\label{Related works}
\subsection{Scene Recognition}
Scene recognition is a pivotal research topic within the field of scene understanding \cite{xie2020scene}. Recently, several deep neural network architectures \cite{he2016deepr5,simonyan2014veryr6,howard2019searchingr7,ma2018shufflenetr8,mehta2021mobilevitr9,liu2021swinr10,dosovitskiy2020imager11,liu2022convnetr12} have been used to facilitate the development of computer vision. Accordingly, some methods attempt to extract deep features for scene recognition. Xie et al. \cite{xie2015hybridr13} propose to combine CNN with dictionary-based representations for scene recognition. Lin et al. \cite{xie2019hierarchicalr14} propose a hierarchical coding algorithm to transform convolutional features into the final image representation for scene recognition. However, while neural network architectures have triumphed in classical domains like image classification, their efficacy in scene recognition often falls short. This disparity is mainly due to the high inter-class similarity resulting from the existing objects across scenes. Therefore, numerous methods utilize internal object information for scene recognition \cite{herranz2016scener15,song2017multir16,lin2022scener47}. 

Specifically, Herranz et al. \cite{herranz2016scener15} propose a method for enhancing scene classification by dividing scene images into patches of different scales and merging them. Similarly, Song et al. \cite{song2017multir16} propose using neural networks to create discriminative patch representations and build a hierarchical architecture for scene recognition. These approaches aim to extract patch-level representations from dense grids to guide scene recognition. However, using dense grids can lead to semantic ambiguity, such as a complete object being distributed across multiple patches or multiple objects residing within a single patch. Considering this limitation, some other methods \cite{sun2018fusingr17,cheng2018scener19,song2019imager20,li2021place} use object detection to extract more certain object regions from the scene. For instance, \cite{sun2018fusingr17} employs detection techniques to extract a high-level deep representation of objects, which is combined with the backbone to create comprehensive representations for scene recognition. Song et al. \cite{song2019imager20} propose a framework that captures object-to-object relations for representing images in scene recognition. Furthermore, some researchers have recognized the background's crucial role in scene classification. Consequently, several approaches \cite{lopez2020semanticr21,zeng2020amorphousr22,song2023srrmr45} incorporate semantic segmentation to extract a scene's foreground and background. SAS-Net \cite{lopez2020semanticr21} enhances scene classification by weighting feature representations based on semantic features obtained from semantic segmentation score tensors, allowing the network to focus on discriminative regions within images. ARG-Net \cite{zeng2020amorphousr22} utilizes semantic segmentation to segment regions within the scene. It then combines the feature maps derived from the backbone to establish context between regional features. 

However, either multi-branching (patch-level) or object-assisted approaches (object detection, semantic segmentation) impose heavy computational burdens on the model,  rendering it unsuitable for practical deployment on resource-constrained edge devices. In contrast, the proposed similarity prototype improves the discriminative ability of networks by introducing semantic prior knowledge during training. This approach enables the achievement of higher recognition accuracy when applying the trained model in real-world scenarios. In this context, the model can rely solely on the exceptional feature extraction capabilities of the backbone network itself, without the need to shoulder additional computational burdens like semantic segmentation.

\subsection{Label Softening}
In recent years, it has been widely acknowledged that training DNNs with hard labels tends to lead to overfitting \cite{muller2019doesr25}. Consequently, label softening has been utilized in various applications. One such technique is Label Smoothing Regularization (LSR) \cite{szegedy2016rethinkingr23}, which involves taking an average between the hard labels and a uniform distribution over labels to soften labels. This approach prevents the network from rapidly overfitting due to overconfidence. Another method, Bootstrapping \cite{reed2014trainingr24} introduces two label softening methods, Bootsoft and Boothard, to mitigate the negative impact of noisy labels. Bootsoft applies predictive distributions to smooth labels, while Boothard uses predictive categories to soften labels. The authors in \cite{li2019reconstructionr26} propose to embed images and labels into a latent space to capture their inherent relationships. By doing so, they leverage this latent space to regularize the network and improve classification performance. Online Label Smoothing (OLS) \cite{zhang2021delvingr27} suggests a strategy where soft labels are generated based on the model’s prediction statistics for the target category. This online label smoothing further enhances the regularization ability of soft labels. Similarly, label smoothing is introduced in \cite{gao2022labelr28} within a hybrid loss function to assist with variable hyperparameters and improve the accuracy of model loss calculations.

Unlike the mentioned approaches above, our main objective in label softening is to embed the inter-class correlation in the proposed similarity prototype into the labels. This embedding guides the network to fit in a more discriminative direction during training. Of course, due to the weakened confidence of the target classes, our softened labels still have the ability to mitigate the overfitting of the network.

\subsection{Deep Metric Learning}
Deep Metric Learning (DML) aims to use deep networks to map data into a nonlinear embedding space, where similar data are close together, and dissimilar data are far apart \cite{kaya2019deepr31}. In general, DML can be divided into two main categories. The former is the Siamese network (contrastive loss) \cite{chopra2005learningr29}, a method based on one pair of samples at a time, which greedily increases the similarity of positive pairs and reduces negative pairs. The latter is the Triplet network (triplet loss) \cite{hoffer2015deepr30}, which pulls the anchor sample closer to the positive sample than the negative sample by a fixed margin. Furthermore, PSDML \cite{ni2017finer32} proposes the quadruple loss, which sets boundaries for anchor-positive and anchor-negative pairs, respectively. N-pair loss \cite{sohn2016improvedr33} and Structured loss \cite{oh2016deepr34} propose to explore the relationship between multiple pairs of positive and negative samples simultaneously within a training batch, utilizing more information to achieve a faster fit. However, none of the above optimization approaches are flexible enough as they apply the same distance threshold to all pairs, regardless of the potential variance in their similarities and dissimilarities. With this in mind, several methods \cite{gonzalez2022guidedr36,zhang2023graphr37} attempt to adaptively distinguish similarity differences based on the network optimization state or the degree of difference between sample pairs. Nonetheless, these methods rely on the discriminative capacity of the network, thus limiting their effectiveness. In contrast, our approach utilizes a statistically derived similarity prototype as prior knowledge, allowing us to assign suitable boundaries to each class-level sample pair, thereby improving the metric's performance.

\section{Proposed method}
\label{Proposed method}
\subsection{Motivation and overall description}
Different from previous semantic-guided methods, we choose to use semantic knowledge to help networks achieve higher scene recognition performance without additional computational costs in practice. This means that computationally intensive and time-consuming object extraction tasks are no longer performed in real-world applications of the network. For this reason, instead of focusing on exploring the intra-scene object region context, we emphasize discovering the intrinsic similarity among scene classes with the help of semantic knowledge. As we introduced in the previous section, the inter-scene similarity can help the network to find a correct fitting direction during training, i.e., to devote more effort to discriminating similar classes.

Specifically, we first derive semantic representations for each scene category using semantic segmentation techniques, training data, and statistics. These semantic representations are then used to construct a similarity prototype, which contains detailed inter-class similarity knowledge. Furthermore, we propose two approaches for leveraging the similarity prototype to assist in network training: Gradient Label Softening (GLS) and Batch-level Contrastive Loss (BCL). In this way, we make the trained network achieve higher performance without adding any additional network parameters or computational costs in practice.

Such a plug-and-play technique can be easily incorporated into most scene representation learning approaches, all it takes is just a few lines of code with minimal computational overhead during training.

\subsection{Similarity Prototype}
\label{subsec: Similarity Prototype}
\begin{figure*}[htbp]
    \centering
    \includegraphics[width=12cm]{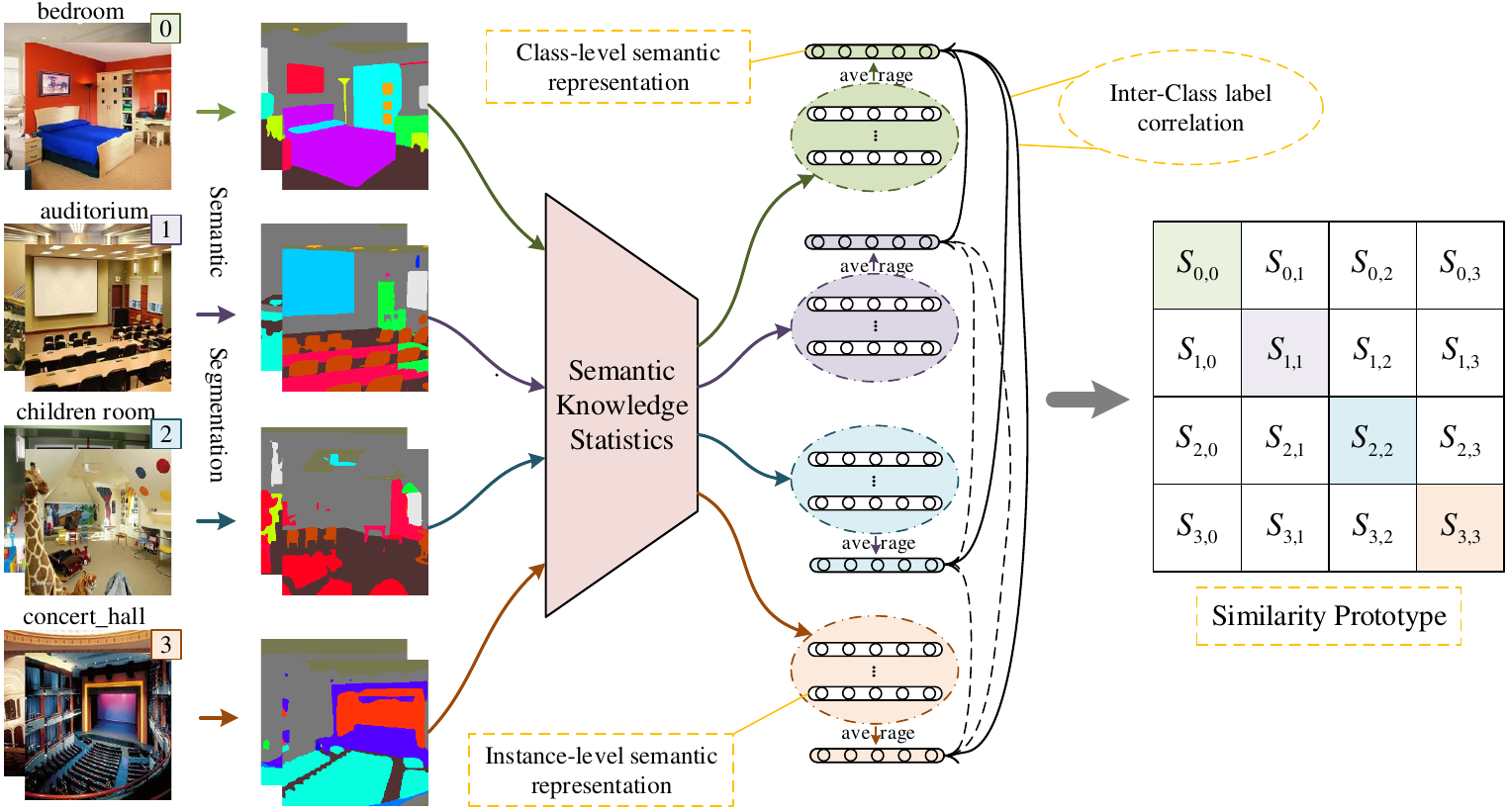}
    \caption{An illustration of the overall process of making a similarity prototype for a scene dataset. Different scene categories are represented by different colored blocks. The derived similarity prototype is denoted as a matrix with dimensions equal to the number of scene categories, with all diagonal elements being 1. $S_{i,j}$ denotes the label correlation between two scene classes.}
    \label{Fig2}
\end{figure*}

The reasoning process for the similarity prototype can be outlined in two steps: semantic representation and label correlation. As shown in Fig. \ref{Fig2}, we initially use the semantic knowledge to derive a semantic representation for each scene class. These class-level semantic representations serve as the basis for investigating inter-class label correlations, which are subsequently embedded into our similarity prototype.

\subsubsection{Class-level semantic representations}

Given a scene dataset, we denote the training data of it by ${D_T} = \left\{ {\left( {{X_c},{y_c}} \right)} \right\}_{c = 1}^C$, where $X_c$ represents all training instances of the ${c_{th}}$ scene class and ${y_c} \in \left\{ {1,2,..., C} \right\}$ is the corresponding label of $X_c$. For each scene class $X_c$, we further denote it as ${X_c} = \left\{ {x_c^n,{y_c}} \right\}_{n = 1}^N$, where $x_c^n \in {\mathbb{R}^{W \times H \times 3}}$ represents the $n_{th}$ training instance of $c_{th}$ scene class. To obtain the class-level semantic representation, we integrate all instance-level representations of ${X_c}$.

The pseudocode for generating class-level semantic representations is presented in Algorithm \ref{alg:alg1}. For each scene instance $x_c^n$, the first step is to feed it into a semantic segmentation network to gain its semantic segmentation label map $M_c^n \in {\mathbb{R}^{W \times H}}$. $M_c^n$ represents the object semantic label of each pixel in $x_c^n$ and can be denoted as $M_{c}^{n}=\left\{M_{c}^{n}(w, h)\right\}_{\substack{1 \leq w \leq W \\ 1 \leq h \leq H}}$. $M_c^n$ is then used to derive the instance-level semantic representation $s_c^n \in {\mathbb{R}^L}$ of $x_c^n$. $L$ denotes the number of semantic categories ($L=150$). Concretely, for any $l \in \left[ {1,L} \right]$ , if there exists $\left( {w,h} \right)$ such that $M_c^n\left( {w,h} \right) = l$, then $s_c^n\left( l \right) = 1$; conversely, if $M_c^n\left( {w,h} \right) \ne l$ for any $\left( {w,h} \right)$, then $s_c^n\left( l \right) = 0$. The instance-level semantic representation $s_c^n$ of $x_c^n$ is obtained by performing the above procedure $L$ times.

\begin{algorithm}[htb]
\caption{Class-level semantic representations algorithm}\label{alg:alg1}
\textbf{Input}: training instances $x_c^n, c=1,...C; n=1,...,N$\\
\raggedright
\textbf{Output}: class-level semantic representation ${S_c} \in {\mathbb{R}^L}$
\begin{algorithmic} 
\FOR{$c=1$ to $C$}
    \FOR{$n=1$ to $N$}
    \STATE Let instance-level semantic representation ${s_c^n} \in {\mathbb{R}^L}$ be a zero vector.
    \STATE $M_c^n = \text{SemanticSegmentation}(x_c^n)$

        \FOR{$l=1$ to $L$}
            \IF {$l$ in $M_c^n$}
                \STATE $s_c^n(l) = 1$
            \ENDIF
        \ENDFOR
    \ENDFOR
    \STATE ${S_c} = \frac{1}{N}\sum\limits_{n = 1}^N {s_c^n}$
\ENDFOR
\end{algorithmic}
\end{algorithm}

By repeating the above process $N$ times, we can obtain all the instance-level semantic representations in the $c_{th}$ scene class $X_c$, which are used to compute the class-level semantic representation ${S_c} \in {\mathbb{R}^L}$ of $X_c$:
\begin{equation}
\label{eq1}
{S_c} = \frac{1}{N}\sum\limits_{n = 1}^N {s_c^n} 
\end{equation}
where $N$ represents the number of training instances of the $c_{th}$ scene class.

After looping through the above process $C$ times, we obtain all the class-level semantic representations, which are then used to derive inter-class label correlation.

\subsubsection{Inter-Class label correlation}
Based on the obtained $C$ class-level semantic representations, we measure the label correlation between different scene classes from two perspectives: Cosine similarity and Euclidean distance. Let us use $S_{i,j}$ to represent the label correlation between the $i_{th}$ and $j_{th}$ scene class.

\paragraph{Cosine similarity-based label correlation}

The cosine similarity captures the similarity between two vectors by computing the cosine of the angle between them. In our case, it is used to quantify the label correlation between two semantic representations, which can be calculated using the following formula:
\begin{equation}
\label{eq2}
S_{i,j}=\text { CosineSimilarity }\left(S_{i}, S_{j}\right)=\frac{S_{i} \cdot S_{j}}{\left\| {S_i}\vphantom{S_j} \right\| \left\| {S_j} \right\|}
\end{equation}
where ${S_i} \cdot {S_j}$ denotes the inner product of $S_i$ and $S_j$, and $\left\| {S_i}\vphantom{S_j} \right\|$, $\left\| {S_j} \right\|$ denote their respective Euclidean norms.

$S_{i,j}$ reflects the label correlation between the $i_{th}$ and $j_{th}$ scene class. A higher value of $S_{i,j}$ indicates a stronger correlation, implying the two classes are more similar to each other and more challenging to distinguish. Conversely, a lower value suggests a weaker correlation, indicating that they can be easily discriminated.

\paragraph{Euclidean distance-based label correlation} 

In addition to cosine similarity, we employ Euclidean distance as a measure of inter-class label correlation. The Euclidean distance quantifies the dissimilarity between two semantic representations by computing the geometric distance between them, which can be formulated as:
\begin{equation}
\label{eq3}
EuclideanDistance({S_i},{S_j}) = \sqrt {\sum\limits_{l = 1}^L {{{({S_{il}} - {S_{jl}})}^2}} } 
\end{equation}
where $L=150$ represents the dimensionality of the semantic representation $S_i$ and $S_j$. $S_{il}$ and $S_{jl}$ denote the values of $S_i$ and $S_j$ respectively in the $l$th dimension.

Eq. \ref{eq3} measures the dissimilarity between semantic representations. To ensure a positive relevance between the label correlation and inter-class similarity, we derive the label correlation $S_{i,j}$ by transforming the Euclidean distance using an exponential function:
\begin{equation}
\label{eq4}
{S_{i,j}} = \exp \left( { - EuclideanDistance\left( {{S_i},{S_j}} \right)} \right) 
\end{equation}

By applying this transformation, a higher value of $S_{i,j}$ indicates a stronger correlation, while a lower value suggests a weaker correlation.

\begin{figure}[htbp]
    \centering
    \includegraphics[width=8cm]{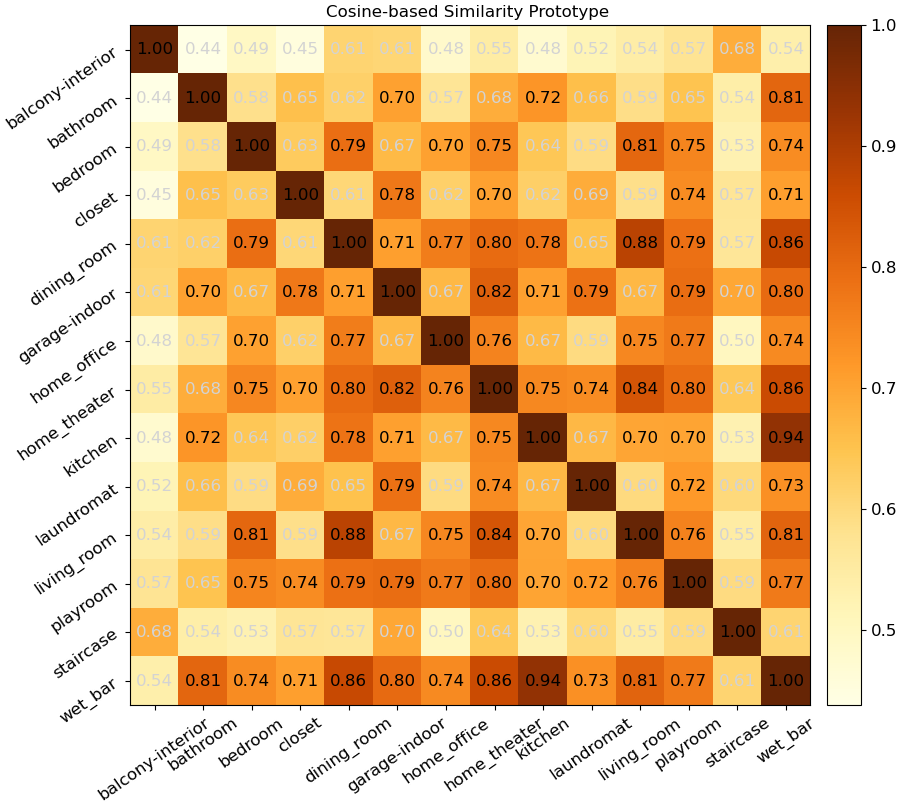}
    \caption{An example of the cosine-based similarity prototype for the Places365-14 dataset. Inter-class label correlations are quantified in the similarity prototype. Darker colors indicate stronger similarity between corresponding scene categories; lighter colors indicate weaker similarity between scene categories.}
    \label{fig3}
\end{figure}

Based on these two measurements, we can construct two similarity prototypes $S \in {\mathbb{R}^{C \times C}}$, in which the value of $S_{i,j}$ represents the label correlation between the $i_{th}$ and $j_{th}$ scene class. Fig. \ref{fig3} illustrates the cosine-based similarity prototype for the Places365-14 dataset. It can be observed that “bedroom” exhibits a closer relationship with “living room” than with “kitchen,” while “kitchen” shows a stronger relationship with “dining room” compared to “bedroom.” These findings are consistent with common sense and are sufficient as prior knowledge.

We have successfully embedded the similarity correlation between scene classes into the similarity prototype. In the following, we present two perspectives on using the similarity prototype to assist model training without adding any network parameter.

\subsection{Similarity prototype-based Gradient Label Softening}
In this section, we propose to utilize the similarity prototype to assist model training from a label-softening perspective. Unlike Label Smoothing Regularization (LSR) \cite{szegedy2016rethinkingr23} that only softens labels with human-defined smoothing values, with similarity prototypes providing prior knowledge, we quantify the degree of label softening in terms of the similarity among different scene classes.

To be concrete, given a dataset with $C$ scene classes, we can reason about its similarity prototype $S \in {\mathbb{R}^{C \times C}}$ (refer to Section \ref{subsec: Similarity Prototype} for details). We first normalize the row vectors of $S$ to obtain $S_{norm}$ in usage soft label form:
\begin{equation}
\label{eq5}
{S_{norm}} = S \oslash \left( {S \cdot {E_{C \times C}}} \right)
\end{equation}

where $\oslash$ represents element-level division, ${E_{C \times C}}$ represents the All-ones matrix in $C \times C$ dimensions.

However, there are differences in the target category confidence of the individual soft labels in $S_{norm}$, which potentially leads to the class imbalance in model training. To solve this, we propose a further refinement to $S_{norm}$, ensuring that each soft label has the same target category confidence. We achieve this by modifying the diagonal elements of $S$. 

Specifically, given ${S_{c,c}}$, its corresponding target category confidence $\sigma$ in the soft label can be calculated by:
\begin{equation}
\label{eq6}
\sigma  = \frac{{{S_{c,c}}}}{{\sum\nolimits_{i = 1}^C {{S_{i,c}}} }} = \frac{{{S_{c,c}}}}{{{S_{c,c}} + \sum\nolimits_{i = 1,i \ne c}^C {{S_{i,c}}} }}
\end{equation}

After conversion, given the target confidence $\sigma '$, its corresponding ${S_{c,c}}$ should be assigned as:
\begin{equation}
\label{eq7}
{S_{c,c}} = \frac{{\sigma '}}{{1 - \sigma '}}\sum\nolimits_{i = 1,i \ne c}^C {{S_{i,c}}}
\end{equation}

Next, we take the maximum target category confidence in $S_{norm}$ as the unity confidence, and update $S$ to obtain ${S^1}$:
\begin{equation}
\label{eq8}
\begin{gathered}
  \sigma'_0 = \max \left( {{S_{norm}}} \right) \hfill \\
  S_{i,j}^1 = \left\{ {\begin{array}{*{20}{c}}
  {\frac{{\sigma'_0}}{{1 - \sigma'_0}}\sum\nolimits_{i = 1,i \ne c}^C {{S_{i,j}}} }&{i = j} \\ 
  {{S_{i,j}}}&{i \ne j} 
\end{array}} \right. \hfill \\ 
\end{gathered} 
\end{equation}

Then, using Eq. \ref{eq5} again, we can obtain the available normalized matrix $S_{norm}^1$. In $S_{norm}^1$, different class labels possess the same confidence for the target category, while maintaining a reasonable confidence difference between the target category and non-target categories.

\begin{algorithm}[htb]
\caption{Gradient Label Softening algorithm}\label{alg:alg2}
\textbf{Input}: similarity prototype $S$\\
\textbf{Output}: softened label
\begin{algorithmic} 
\STATE \COMMENT{$STEP$: The number of epochs using soft labels.}

\FOR{$epoch=1$ to $End\_Epoch$}
\STATE $\sigma' = \sigma'_0 + \left( {0.99 - \sigma'_0} \right) \cdot \frac{{epoch - 1}}{{STEP}}$
\STATE ${S^1} = S.copy()$
    \FOR{$c=1$ to $C$}
        \STATE $S_{c,c}^1 = \frac{{\sigma'}}{{1 - \sigma'}}\sum\nolimits_{i = 1,i \ne c}^C {{S_{i,c}}}$        
    \ENDFOR
    \STATE $S_{norm}^1 = {S^1} \oslash \left( {{S^1} \cdot {E_{C \times C}}} \right)$ \# soft label
    \IF{$\sigma' > 0.99$} 
        \STATE $S_{norm}^1 = {I_{C \times C}}$ \# hard label
    \ENDIF
    \STATE \textbf{return} $S_{norm}^1$
\ENDFOR
\end{algorithmic}
\end{algorithm}

Since the inference process of the similarity prototype overlooks specific details like colors within scenes, intuitively, the trained model's ability to differentiate between various scene categories should surpass that of the pure knowledge statistics (similarity prototype). Consequently, instead of constraining the target category confidence $\sigma'$ to $\sigma'_0$, we gradually boost the target category confidence during training, thereby driving the network to progressively improve its discriminative ability in the appropriate direction. With the increase of training epoch, the confidence $\sigma'$ of the target category can be formulated as:

\begin{equation}
\label{eq_value}
\sigma' = \sigma'_0 + \left( {0.99 - \sigma'_0} \right) \cdot \frac{{epoch - 1}}{{STEP}}
\end{equation}
where $STEP$ represents the number of epochs using soft labels. We use $\sigma'$ in place of $\sigma'_0$ in Eq. \ref{eq8} and reapply Eq. \ref{eq8} and Eq. \ref{eq5} to update class labels during training, as outlined in Algorithm 2.

\begin{figure*}[htbp]
    \centering
    \includegraphics[width=12cm]{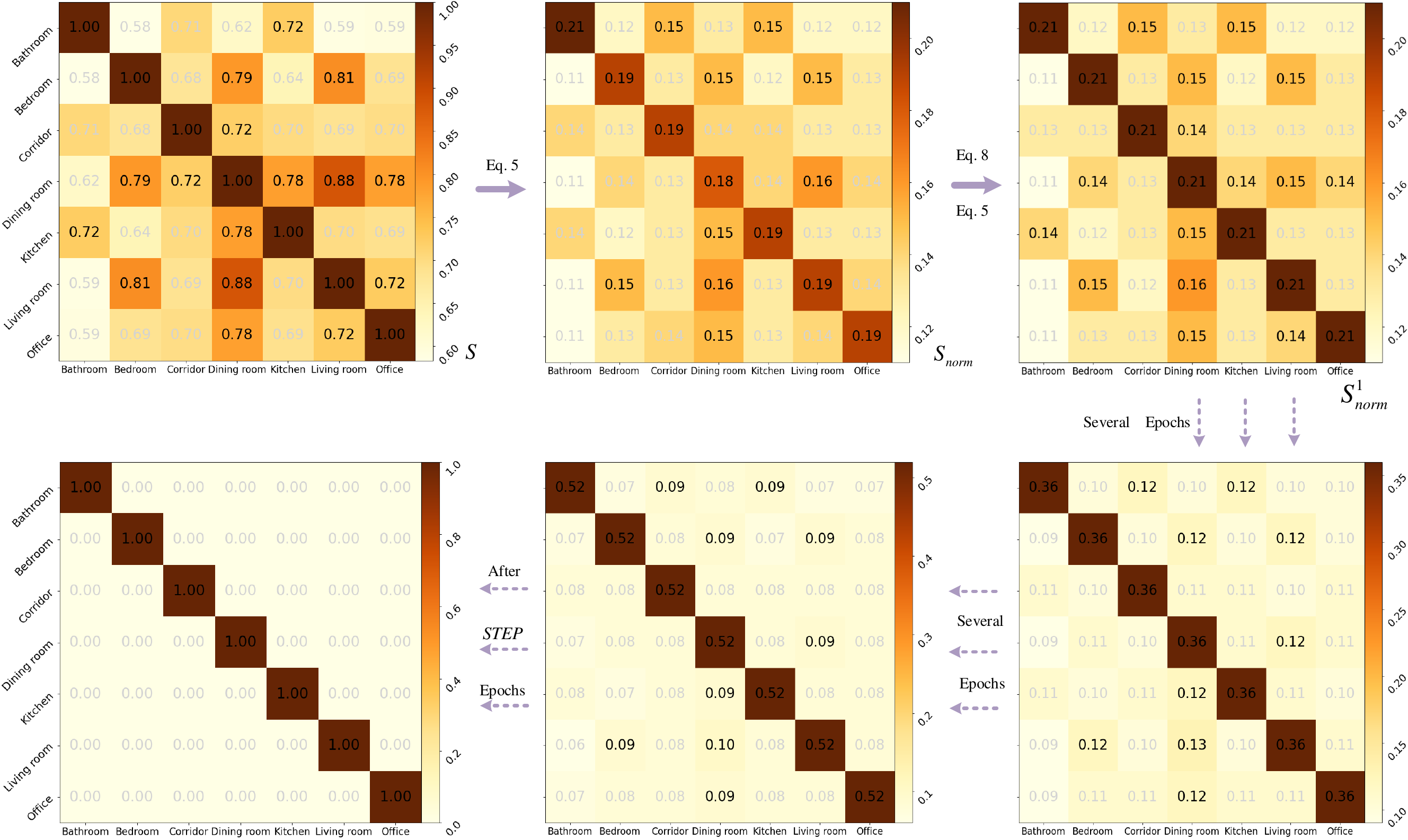}
    \caption{Taking the cosine-based similarity prototype for the Places365-7 as an example, $S_{norm}^1$ varies with epoch growing. As the target category confidence increases, attention towards the non-target category gradually decreases. In the process, $S_{norm}^1$ still maintains the difference in attention towards different non-target categories. Once $STEP$ epochs have passed, all class labels are converted to a hard label.}
    \label{Fig4}
\end{figure*}

In the above process, we maintain the confidence difference among scene categories based on the similarity prototype, ensuring that the network fully uses semantic prior knowledge. After $STEP$ epochs (when $\sigma' > 0.99$), all the class labels are converted into a hard label. An example of the changes in $S_{norm}^1$ across epochs is shown in Fig. \ref{Fig4}.

With the change of epoch, the soft labels supplied by $S_{norm}^1$ will be used as fitted endpoints for model training along with the cross-entropy loss function, as follows:

\begin{equation}
\label{eq9}
{\mathcal{L}_{CE}} =  - \frac{1}{B}\sum\nolimits_{i = 1}^B {\sum\nolimits_{j = 1}^C {\left( {S_{norm}^1\left[ {j,target[i]} \right] \cdot \log \left( {p\left[ {i,j} \right]} \right)} \right)} }
\end{equation}
where $B$ represents the number of images in a minibatch, $p$ represents the prediction probabilities output by the network, and $target$ represents the corresponding ground truth of the mini-batch.

\subsection{Similarity prototype-based Batch-level Contrastive Loss}
\label{subsec: BCL}

\begin{figure}[htbp]
    \centering
    \includegraphics[width=10cm]{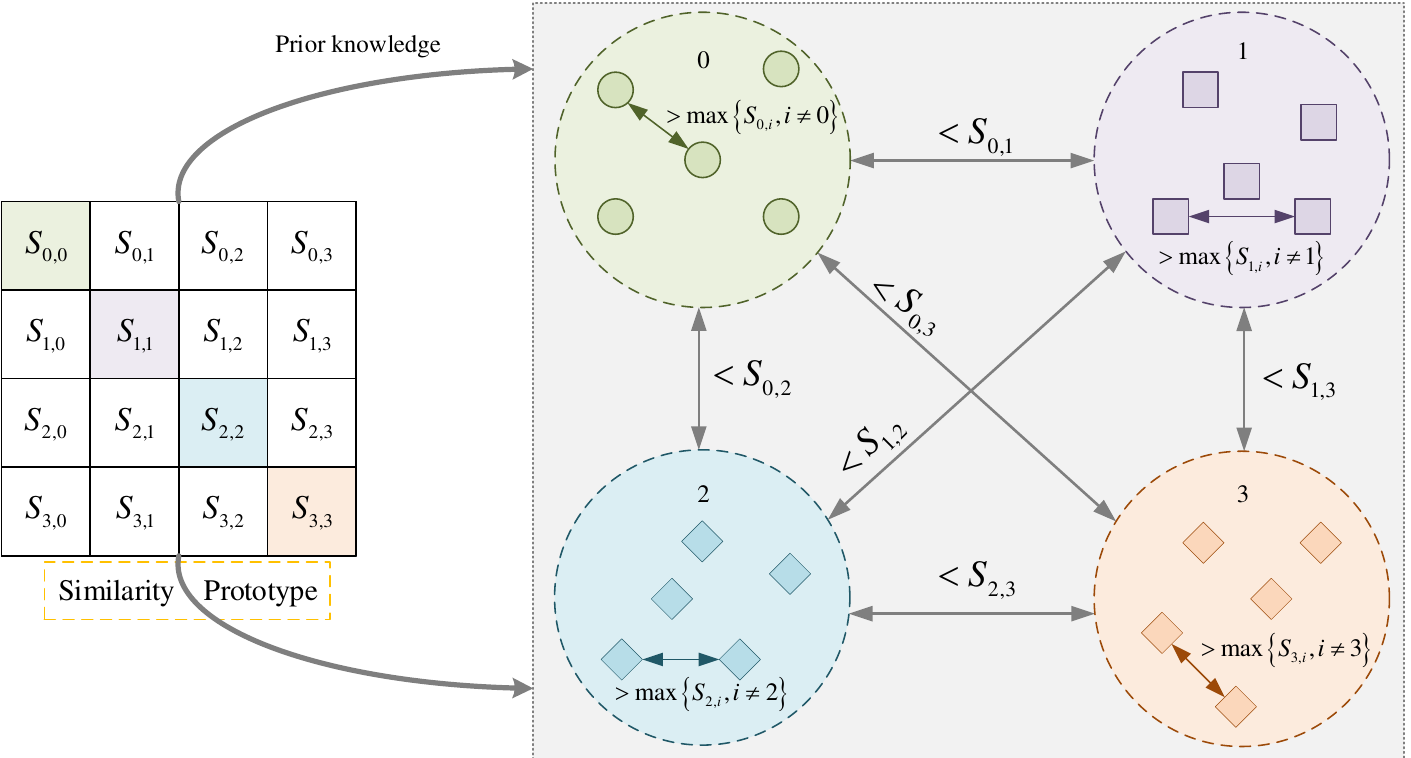}
    \caption{An illustration of the operation of the proposed Contrastive Loss function guided by the proposed Similarity Prototype. Different scene categories are represented by different colored blocks. $S_{i,j}$ denotes the label correlation between two scene classes.}
    \label{fig5}
\end{figure}

In this section, we use the similarity prototype to support model training from a metric learning perspective (i.e., contrastive loss). Contrastive loss aims to optimize the feature space by maximizing similarity within the same class and minimizing similarity between different classes. By integrating the similarity prototype into contrastive loss, we eliminate the need for manual threshold definition and instead establish more suitable boundaries based on statistical similarity. Fig. \ref{fig5} visually represents the batch-level contrastive loss operation. Let $x_i^m$ and $x_j^n$ be a pair of inputs, the contrastive loss ${\mathcal{L}_{contrastive}}$ can be formulated as:
\begin{equation}
\label{eq10}
\begin{split}
{\mathcal{L}_{\text{contrastive}}} = &(1 - \gamma) \max\{0, p_{x_i^m,x_j^n} - S_{i,j}\} + \gamma \max\{0, \max\{S_{i,j}, j \ne i\} - p_{x_i^m,x_j^n}\}
\end{split}
\end{equation}
where ${p_{x_i^m,x_j^n}}$ represents the feature similarity between $x_i^m$ and $x_j^n$. If a pair of inputs is from the same class, the value of $\gamma$ is 1; otherwise, its value is 0.

To ensure the plug-and-play nature, inspired by N-pair loss \cite{sohn2016improvedr33} and Structured loss \cite{oh2016deepr34}, we compute contrastive loss based on mini-batch. Specifically, assume that the input to the network is a mini-batch of $B$ randomly sampled scene images. Denote the output prediction of the network as $p \in {\mathbb{R}^{B \times C}}$ and the ground truth as $target \in {\mathbb{R}^B}$. Using $target$ as the row index of similarity prototype $S$, we can calculate the similarity prototype variant ${S_{batch}} \in {\mathbb{R}^{B \times B}}$ among the $B$ samples in $target$ via
\begin{equation}
\label{eq11}
{S_{batch}} = S\left[ {target} \right] \cdot S{\left[ {target} \right]^T}
\end{equation}

Next, we can calculate the similarity matrix ${p_{matrix}} \in {\mathbb{R}^{B \times B}}$ among the $B$ samples within the network's prediction $p$ using cosine similarity or Euclidean distance.

Since the inference process of the similarity prototype overlooks details like colors with scenes, the discriminability of the trained model should surpass the statistically derived similarity prototype. Therefore, the similarity between scene classes predicted by the network should be lower than the derived inter-class similarity. Accordingly, an inter-class contrastive loss matrix ${\mathcal{L}_{inter\_matrix}}$ can be designed via
\begin{equation}
\label{eq12}
{\mathcal{L}_{inter\_matrix}} = {p_{matrix}} - {S_{batch}}
\end{equation}

Note that we only need to consider the anomalous case, so the inter-class contrastive loss ${\mathcal{L}_{inter}}$ is formulated by
\begin{equation}
\label{eq13}
{\mathcal{L}_{inter}} = Mean\left( {\max \left( {{\mathcal{L}_{inter\_matrix}},0} \right)} \right)
\end{equation}

However, the above approach only considers samples from different scene classes, failing to leverage the feature information of samples from the same class within the minibatch. The reason is that the similarity between samples within a class cannot be effectively measured using the similarity prototype, and it is unrealistic to achieve a similarity of "1" between different samples of the same class. However, in the scene classification task, we only need to ensure that the model outputs a higher similarity between samples of the same category than its similarity to other categories (as shown in Fig. \ref{fig5}). Therefore, we improve the contrastive loss by a layer of intra-class contrast measures to better utilize the feature information of samples from the same class.

To compute the intra-class contrastive loss, we first generate a variant of the similarity prototype $S$: the self-similarity matrix $S''$. This matrix solely focuses on intra-class similarity. It sets the intra-class similarity threshold to the maximum value obtained from the similarity between the scene class and other scene classes:
\begin{equation}
\label{eq14}
{S''_{i,j}} = \left\{ {\begin{array}{*{20}{c}}
  {\max \left\{ {{S_{i,k}},i \ne k} \right\}}&{i = j} \\ 
  0&{i \ne j} 
\end{array}} \right.
\end{equation}

Similarly, we can obtain the intra-class similarity prototype variant ${S''_{batch}} \in {\mathbb{R}^{B \times B}}$ among the $B$ samples in $target$ via Eq. \ref{eq11}. Then, the intra-class contrastive loss matrix ${\mathcal{L}_{intra}}$ can be designed via
\begin{equation}
\label{eq15}
\begin{gathered}
  {\mathcal{L}_{intra\_matrix}} = {{S''}_{batch}} - {p_{matrix}} \hfill \\
  {\mathcal{L}_{intra}} = Mean\left( {\max \left( {{\mathcal{L}_{intra\_matrix}},0} \right)} \right) \hfill \\ 
\end{gathered}
\end{equation}

As for whether the measure of intra-class contrastive loss is effective, we conduct experiments for comparative discussion in Section IV.

Of course, the contrastive loss is still secondary after all, and the final loss is generated by the combination of the cross-entropy loss and the contrastive loss:
\begin{equation}
\label{eq16}
\mathcal{L} =  - \frac{1}{B}\sum\nolimits_{i = 1}^B {\log \left( {\frac{{\exp \left( {{p_{i,targe{t_i}}}} \right)}}{{\sum\nolimits_{j = 1}^C {\exp \left( {{p_{i,j}}} \right)} }}} \right)}  + \left( {{\mathcal{L}_{inter}} + {\mathcal{L}_{intra}}} \right)
\end{equation}

\section{Experiments}
\label{Experiments}
In this section, we aim to evaluate the effectiveness of the proposed similarity prototype. We conduct separate evaluations of the proposed Gradient Label Softening and Batch-level Contrastive Loss methods on the MIT-67 \cite{quattoni2009recognizingr38} and SUN397 \cite{xiao2010sunr39} datasets. Subsequently, we apply these methods to two simplified versions of the Places365 \cite{zhou2017placesr40} datasets to visualize their functionality. Finally, we compare our approach with existing state-of-the-art methods.

\subsection{Implementation Details}
\textbf{Semantic Segmentation}: Vision Transformer Adapter \cite{chen2023visionr41} that is pretrained on the ADE20K dataset \cite{zhou2017scener42} is used as the semantic segmentation network. The network outputs a label map $M$, which has the same size as the input image. Each pixel $\left( {w,h} \right)$ in $M$  is assigned a value ${M_{wh}}$, representing the semantic label of the corresponding pixel in the input image.

\textbf{Hyperparameters}: The proposed similarity prototype and its derivative strategies can be easily integrated into various models. We conduct a series of experiments to demonstrate the effectiveness and generalization of our methods using eight commonly used pretrained models: ResNet50-IN1k \cite{he2016deepr5}, VGG16-IN1k \cite{simonyan2014veryr6}, MobileNetV3-IN1k \cite{howard2019searchingr7}, ShuffleNetV2-IN1k \cite{ma2018shufflenetr8}, MobileVIT\_S-IN1k \cite{mehta2021mobilevitr9}, ConvNextBase-IN22k \cite{liu2022convnetr12}, VITBase-IN22k \cite{dosovitskiy2020imager11}, and SwinTransformerBase-IN22k \cite{liu2021swinr10}. The suffix “IN1k” indicates that the model is pretrained on the ImageNet 1k dataset \cite{russakovsky2015imagenetr43}, while “IN22k” indicates pretrained on the ImageNet 22k dataset \cite{russakovsky2015imagenetr43}.

We train all models using Adam optimizer \cite{kingma2014adamr44}. For the validation on the MIT-67 and SUN397 datasets, we set the initial learning rate of the last fully connected layer to 0.001, and the learning rate of the other pretrained layers to 0.00001 (decayed by a factor of 0.1 at the 10th, 15th, and 20th epochs). The batch size is set to 32, and the weight decay to 0.00001. We train the models for a total of 100 epochs. On the Places365 dataset, due to its larger size and faster convergence, we modify the batch size to 64 and train all models for 30 epochs. When training the models pretrained on ImageNet 22k, considering their larger parameter sizes and faster convergence, we also train them for 30 epochs. For the MobileVIT model, we increase the initial learning rate by a factor of 10 to improve fitting efficiency. All experiments are conducted on a single NVIDIA 3090 GPU using the PyTorch and SAS-Net \cite{lopez2020semanticr21} open-source framework.

\subsection{Datasets}
\textbf{MIT-67} Dataset \cite{quattoni2009recognizingr38} comprises 67 indoor scene classes with a total of 15620 images. Each scene category contains a minimum of 100 images.  In line with the recommendations of \cite{quattoni2009recognizingr38}, each class has 80 images for training and 20 for testing.

\textbf{SUN397} Dataset \cite{xiao2010sunr39} is a comprehensive dataset covering indoor and outdoor scenes. It encompasses 397 scene categories, with 175 indoor and 220 outdoor categories, all comprising at least 100 RGB images. Following the evaluation protocol in the original paper, we randomly select 50 images from each scene class for training and another 50 for testing.

\textbf{Places365} Dataset \cite{zhou2017placesr40} is one of the largest scene-centric datasets, containing approximately 1.8 million training images and 365 scene categories. To visualize the effect of our proposed method, this paper uses a simplified version known as Places365-7 and Places365-14. Places 365-7 consists of seven indoor scenes: Bath, Bedroom, Corridor, Dining Room, Kitchen, Living Room, and Office. Places 365-14 contains 14 indoor scenes: Balcony, Bedroom, Dining Room, Home Office, Kitchen, Living Room, Staircase, Bathroom, Closet, Garage, Home Theater, Laundromat, Playroom, and Wet Bar. For the test set, we use the same setup as the official dataset.

\subsection{Ablation Study}
\subsubsection{Hyperparameter $STEP$ of GLS}
\begin{figure}[htbp]
\centering
\subfloat[]{\includegraphics[width=4cm]{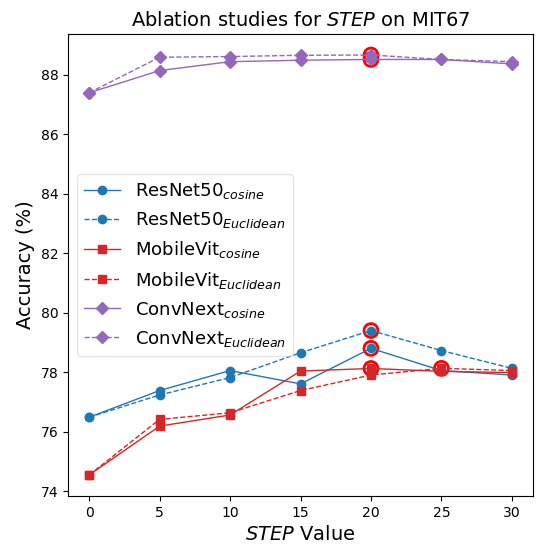}%
\label{fig6_1_case}}
\hfil
\subfloat[]{\includegraphics[width=4cm]{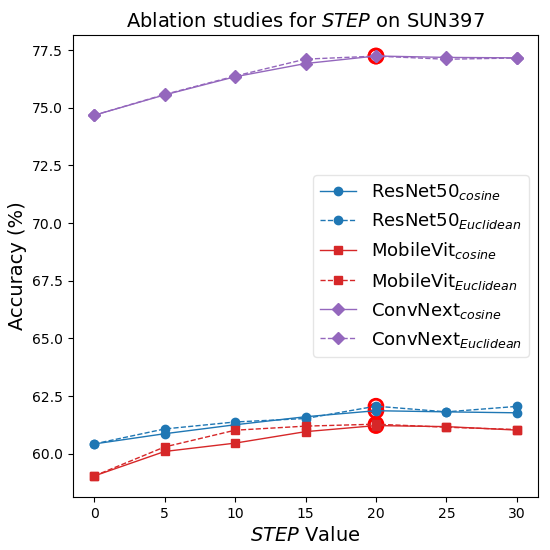}%
\label{fig6_2_case}}
\caption{ Impact of different $STEP$ values on the fitted model accuracy. Note that the nodes circled by hollow bold circles are the highest accuracy points, and $STEP=0$ corresponds to the result when models are trained using hard-label.}
\label{fig6}
\end{figure}

Given the critical role of the hyperparameter $STEP$ in the GLS training strategy, we conduct a series of experiments on the MIT-67 and SUN397 datasets to determine its optimal value. As illustrated in Fig. \ref{fig6}, we evaluate the performance of ResNet and ConvNext models on these datasets for various $STEP$ values. Notably, $STEP=0$ corresponds to the results of models trained using hard-label. By systematically varying the $STEP$ parameter, we observe similar auxiliary effects in terms of cosine similarity-based and Euclidean distance-based similarity prototypes. As we can see: (1) Irrespective of the $STEP$ value, models trained with prior knowledge from similarity prototypes consistently outperform those trained using hard-label, thereby validating the feasibility of embedding the similarity prototype into labels. (2) Our comprehensive experimentation demonstrates that the accuracy of the fitted model gradually increases as the $STEP$ value ranges from 0 to 20, and basically reaches the highest level when reaching 20. Further increasing the $STEP$ value has minimal impact on model accuracy. To ensure consistency, a $STEP$ value of 20 has been chosen for all subsequent model training processes.

\subsubsection{Computation of BCL}
\begin{figure}[htbp]
\centering
\subfloat[]{\includegraphics[width=4.5cm]{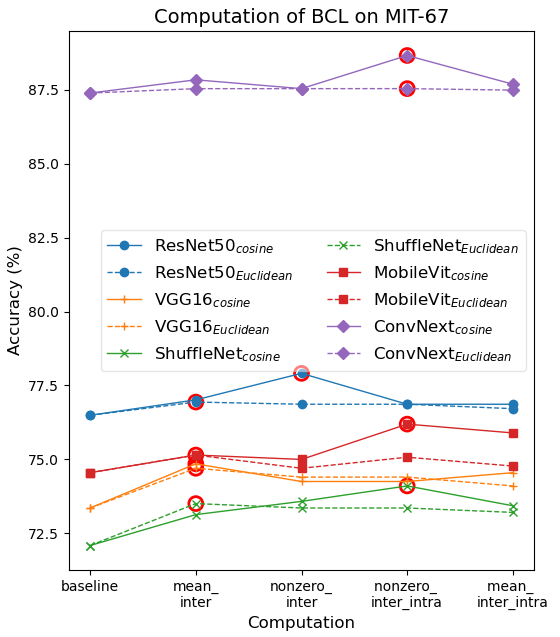}%
\label{fig7_1_case}}
\hfil
\subfloat[]{\includegraphics[width=4.5cm]{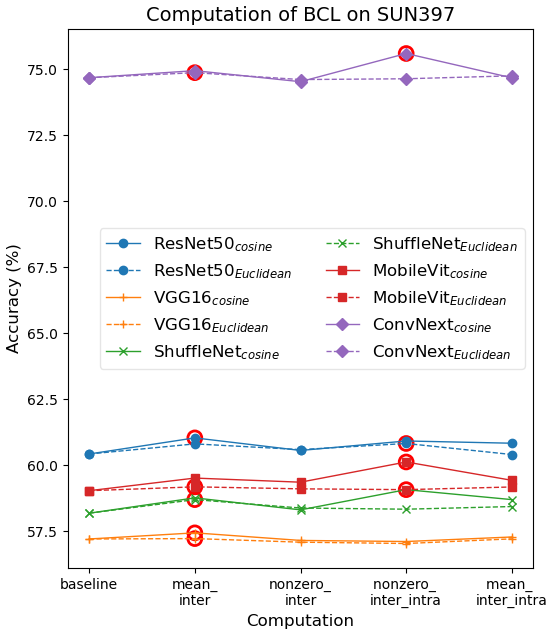}%
\label{fig7_2_case}}
\caption{Performance comparison of contrastive loss-assisted model training under different loss calculation methods. Note that the nodes circled by hollow bold circles are the highest accuracy points.}
\label{fig7}
\end{figure}

Given a contrastive loss matrix (see Section \ref{subsec: BCL} for details), there are two ways for computing the final contrastive loss: global averaging (referred to as "mean") and non-zero averaging (i.e., averaging considering only the number of non-zero values in the matrix, abbreviated as "nonzero"). This section investigates the impact of these computational approaches on model performance and discusses the consideration of intra-class contrastive loss (see Section \ref{subsec: BCL} for details). Comparative experiments were conducted on the MIT-67 and SUN397 datasets using multiple networks, and the results are presented in Fig. \ref{fig7}. Note that for convenience, given the obtained contrastive loss matrix $\mathcal{L}_{inter\_matrix}$ and  $\mathcal{L}_{intra\_matrix}$, we use "mean\_inter" to denote performing global average on $\mathcal{L}_{inter\_matrix}$ to compute final contrastive loss. "mean\_inter\_intra" denotes performing global average on $\mathcal{L}_{inter\_matrix} + \mathcal{L}_{intra\_matrix}$. "nonzero\_inter" denotes performing non-zero average on $\mathcal{L}_{inter\_matrix}$. "nonzero\_inter\_intra" denotes performing non-zero average on $\mathcal{L}_{inter\_matrix} + \mathcal{L}_{intra\_matrix}$.

In Fig. \ref{fig7}, it is evident that using the proposed similarity prototype to assist training from the perspective of contrastive loss can improve the model's accuracy compared to the baseline (i.e., training models by only cross-entropy loss based on hard-label). This outcome validates the feasibility of using the similarity prototype as prior knowledge for assisting training. Additionally, both the similarity prototype based on Euclidean distance and cosine similarity yield favorable outcomes in providing prior knowledge. Considering all the experimental results collectively, in most cases, "mean" (global average) performs better when considering only inter-class contrastive loss, while "non-zero"(non-zero average) yields superior results when both inter-class and intra-class contrastive loss are considered. Consequently, the decision was made to employ these two calculations, namely “mean\_inter” and “nonzero\_inter\_intra,” to assist in training the model.

\subsection{Evaluation of Gradient Label Softening}
This section applies the similarity prototype-based Gradient Label Softening (GLS) strategy to train multiple models. We evaluate GLS using multiple models on the MIT-67 and SUN397 datasets and compared them to the baseline (trained with hard labels) and LSR \cite{szegedy2016rethinkingr23}. The results are presented in Table \ref{tab1}.

It is evident that the proposed GLS significantly improves the scene recognition performance on both lightweight and complex models, demonstrating its robustness across different networks. Specifically, when evaluating on the MIT-67 dataset, applying GLS to ResNet50 yields an accuracy of 79.403\%, surpassing the baseline of 2.91\%. In contrast, traditional LSR shows limited improvements and, in some cases, even negatively affects certain networks. For example, on the SUN397 dataset, ResNet50 with GLS can achieve 62.06\% accuracy, which improves ResNet50 alone by 1.637\% and ResNet50 with LSR by 3.032\%, respectively. It is worth noting that the SUN397 dataset comprises 397 distinct scene classes, further highlighting the suitability and performance of our method on large-scale datasets.

\begin{table*}[htbp]
    \centering
    \caption{Evaluation of Gradient Label Softening (GLS) on MIT-67 and SUN397 datasets. The best result of each row is marked in bold.}
    \resizebox{0.8\textwidth}{!}{
    \begin{tabular}{cllclll}
    \hline
        Dataset & Model & Param & Baseline & LSR\cite{szegedy2016rethinkingr23} & GLS\textsubscript{Cosine Similarity} & GLS\textsubscript{Euclidean Distance} \\ \hline
        MIT-67 & ResNet50  & 25M  & 76.493  & 76.417\textsubscript{\textcolor{red}{-0.076}}  & 78.805  & \textbf{79.403}\textsubscript{\textcolor{red}{+2.910}}  \\ 
        MIT-67 & VGG16  & 138M  & 73.358  & 74.078\textsubscript{\textcolor{red}{+0.72}} & 74.104  & \textbf{75.0}\textsubscript{\textcolor{red}{+1.642}}  \\ 
        MIT-67 & MobileNet  & 5.4M  & 69.179  & 68.507\textsubscript{\textcolor{red}{-0.672}}  & 71.045  & \textbf{71.119}\textsubscript{\textcolor{red}{+1.94}}  \\ 
        MIT-67 & ShuffleNet  & 7M  & 72.090  & 73.209\textsubscript{\textcolor{red}{+1.119}}  & \textbf{74.925}\textsubscript{\textcolor{red}{+2.835}}  & 74.701  \\ 
        MIT-67 & MobileVit  & 6M  & 74.552  & 75.522\textsubscript{\textcolor{red}{+0.97}}  & \textbf{78.134}\textsubscript{\textcolor{red}{+3.582}}  & 77.91  \\ 
        MIT-67 & SwinT  & 87M  & 88.681  & 88.657\textsubscript{\textcolor{red}{-0.024}}  & \textbf{89.03}\textsubscript{\textcolor{red}{+0.349}}  & 88.906  \\ 
        MIT-67 & VIT  & 86M  & 86.194  & 86.642\textsubscript{\textcolor{red}{+0.448}}  & 86.940  & \textbf{87.015}\textsubscript{\textcolor{red}{+0.821}}  \\ 
        MIT-67 & ConvNext  & 89M  & 87.388  & 87.239\textsubscript{\textcolor{red}{-0.149}}  & 88.507  & \textbf{88.657}\textsubscript{\textcolor{red}{+1.269}}  \\ \hline 
        SUN397 & ResNet50  & 25M  & 60.423  & 59.028\textsubscript{\textcolor{red}{-1.395}}  & 61.863  & \textbf{62.060}\textsubscript{\textcolor{red}{+1.637}}  \\ 
        SUN397 & VGG16  & 138M  & 57.204  & 57.950\textsubscript{\textcolor{red}{+0.746}}  & 58.503  & \textbf{58.715}\textsubscript{\textcolor{red}{+1.511}}  \\ 
        SUN397 & MobileNet  & 5.4M  & 53.375  & 53.043\textsubscript{\textcolor{red}{-0.332}}  & \textbf{55.657}\textsubscript{\textcolor{red}{+2.282}}  & 54.625  \\ 
        SUN397 & ShuffleNet  & 7M  & 58.181  & 57.950\textsubscript{\textcolor{red}{-0.231}}  & 59.385  & \textbf{60.08}\textsubscript{\textcolor{red}{+1.899}}  \\ 
        SUN397 & MobileVit  & 6M  & 59.033  & 59.283\textsubscript{\textcolor{red}{+0.25}}  & 61.219  & \textbf{61.28}\textsubscript{\textcolor{red}{+2.247}}  \\ 
        SUN397 & SwinT  & 87M  & 75.063  & 74.826\textsubscript{\textcolor{red}{-0.237}}  & 77.013  & \textbf{77.194}\textsubscript{\textcolor{red}{+2.131}}  \\ 
        SUN397 & VIT  & 86M  & 75.078  & 75.043\textsubscript{\textcolor{red}{-0.035}}  & \textbf{75.758}\textsubscript{\textcolor{red}{+0.68}}  & 75.677  \\ 
        SUN397 & ConvNext  & 89M  & 74.675  & 75.017\textsubscript{\textcolor{red}{+0.342}}   & \textbf{77.274}\textsubscript{\textcolor{red}{+2.599}}  & 77.239  \\ \hline
    \end{tabular}
    }
    \label{tab1}
\end{table*}
Moreover, the benefits of GLS are more pronounced in lightweight models. For example, MobileVit's accuracy on the MIT-67 dataset improved by 3.582\% with GLS. In contrast, for SwinT with a large parameter count, GLS only improves its accuracy by 0.349\%. The reason for this discrepancy is that our approach works mainly by improving the feature discrimination ability of the model itself. SwinT, due to its large number of parameters and advanced network architecture, has performed well (88.681\%) under the baseline strategy. Thus, the potential for improvement is relatively limited. However, when faced with the more complex SUN397 dataset, SwinT has more potential for feature discrimination improvement. In this case, our GLS enables a significant improvement in both lightweight and complex models. This further validates the importance of utilizing semantic similarity knowledge to guide network training.

\subsection{Evaluation of Batch-level Contrastive Loss}
\label{eval_BCL}
In this section, we apply the similarity prototype-based Batch-level Contrastive Loss (BCL) to assist model training. We compare its performance with baseline (i.e., training models by only cross-entropy loss based on hard-label) as well as the traditional Contrastive Loss (CL) strategy \cite{chopra2005learningr29}. In CL, the similarity expectation within the same category is set to "1" and "0" for different categories.  These strategies are evaluated using multiple models on the MIT-67 and SUN397 datasets, and the resulting outcomes are presented in Table \ref{tab2}.

\begin{table*}[!htbp]
    \centering
    \caption{Evaluation of Batch-level Contrastive Loss (BCL) on MIT-67 and SUN397 datasets. The best result of each row is marked in bold.}
    \resizebox{1\textwidth}{!}{ 
    \begin{tabular}{cllcllllc}
    \hline
        Dataset & Model & Param & Baseline & CL\cite{chopra2005learningr29} & \makecell[c]{BCL\textsubscript{Cosine} \\ mean\_inter} & \makecell[c]{BCL\textsubscript{Euclidean} \\ mean\_inter} & \makecell[c]{BCL\textsubscript{Cosine} \\ nonzero\_inter\_intra} & \makecell[c]{BCL\textsubscript{Euclidean} \\ nonzero\_inter\_intra} \\ \hline
        MIT-67 & ResNet50  & 25M & 76.493 & 76.642\textsubscript{\textcolor{red}{+0.149}}  & \textbf{77.014}\textsubscript{\textcolor{red}{+0.521}}  & 76.94 & 76.866 & 76.866  \\ 
        MIT-67 & VGG16  & 138M & 73.358 & 74.327\textsubscript{\textcolor{red}{+0.969}}  & \textbf{74.851}\textsubscript{\textcolor{red}{+1.493}}  & 74.701 & 74.254 & 74.403  \\ 
        MIT-67 & MobileNet  & 5.4M & 69.179 & 68.358\textsubscript{\textcolor{red}{-0.821}}  & \textbf{69.478}\textsubscript{\textcolor{red}{+0.299}}  & 68.433 & 68.582 & 68.433  \\ 
        MIT-67 & ShuffleNet  & 7M & 72.090 & 72.564\textsubscript{\textcolor{red}{+0.474}}  & 73.134  & 73.507 & \textbf{74.104}\textsubscript{\textcolor{red}{+2.014}} & 73.358  \\ 
        MIT-67 & MobileVit  & 6M & 74.552 & 74.776\textsubscript{\textcolor{red}{+0.224}}  & 75.149  & 75.149 & \textbf{76.194}\textsubscript{\textcolor{red}{+1.642}} & 75.075  \\ 
        MIT-67 & SwinT  & 87M & 88.681 & 89.328\textsubscript{\textcolor{red}{+0.671}}  & 89.179  & \textbf{89.701}\textsubscript{\textcolor{red}{+1.02}} & 89.403 & 89.328  \\ 
        MIT-67 & VIT  & 86M & 86.194 & 86.342\textsubscript{\textcolor{red}{+0.148}}  & 86.269  & 86.418 & \textbf{86.493}\textsubscript{\textcolor{red}{+0.299}} & 86.493  \\ 
        MIT-67 & ConvNext  & 89M & 87.388 & 87.462\textsubscript{\textcolor{red}{+0.074}}  & 87.835  & 87.537 & \textbf{88.657}\textsubscript{\textcolor{red}{+1.269}} & 87.537  \\ \hline
        SUN397 & ResNet50  & 25M & 60.423 & 60.635\textsubscript{\textcolor{red}{+0.212}}  & \textbf{61.033}\textsubscript{\textcolor{red}{+0.61}}  & 60.806 & 60.916 & 60.821  \\ 
        SUN397 & VGG16  & 138M & 57.204 & 57.09\textsubscript{\textcolor{red}{-0.114}}  & \textbf{57.436}\textsubscript{\textcolor{red}{+0.232}}  & 57.219 & 57.108 & 57.033  \\ 
        SUN397 & MobileNet  & 5.4M & 53.375 & 53.169\textsubscript{\textcolor{red}{-0.206}}  & 53.270  & 53.713 & \textbf{54.055}\textsubscript{\textcolor{red}{+0.68}} & 54.055 \\ 
        SUN397 & ShuffleNet  & 7M & 58.181 & 58.176\textsubscript{\textcolor{red}{-0.005}}  & 58.761  & 58.695 & \textbf{59.073}\textsubscript{\textcolor{red}{+0.892}} & 58.332  \\ 
        SUN397 & MobileVit  & 6M & 59.033 & 59.264\textsubscript{\textcolor{red}{+0.231}}  & 59.511  & 59.179 & \textbf{60.121}\textsubscript{\textcolor{red}{+1.088}} & 59.073  \\ 
        SUN397 & SwinT  & 87M & 75.063 & 74.7\textsubscript{\textcolor{red}{-0.363}}  & 75.037  & 75.088 & \textbf{75.95}\textsubscript{\textcolor{red}{+0.887}} & 74.877  \\ 
        SUN397 & VIT  & 86M & 75.078 & 75.264\textsubscript{\textcolor{red}{+0.186}}  & 75.416  & 75.314 & \textbf{75.521}\textsubscript{\textcolor{red}{+0.443}} & 75.275  \\ 
        SUN397 & ConvNext  & 89M & 74.675 & 74.632\textsubscript{\textcolor{red}{-0.043}}  & 74.942  & 74.866 & \textbf{75.592}\textsubscript{\textcolor{red}{+0.917}} & 74.635  \\ \hline
    \end{tabular}
    }
    \label{tab2}
\end{table*}

The results in Table \ref{tab2} indicate that, in most cases, BCL yields improved performance over the baseline and generally outperforms the traditional CL strategy. Specifically, when applied to the MIT-67 dataset, BCL-trained ShuffleNet exhibits a performance improvement of 2.014\% compared to the baseline, while the traditional CL strategy only yields a 1.044\% improvement. On the SUN397 dataset, BCL-trained ConvNext shows a performance boost of 0.917\% over the baseline, while the traditional CL results in a 0.043\% decrease. This implies that the traditional CL excessively emphasizes similarity within the same categories and promotes dissimilarity between different categories, consequently impairing recognition performance. These findings further emphasize the feasibility and necessity of incorporating the similarity prototype to provide prior knowledge during training.

In addition, comparing results in Tables \ref{tab2} and \ref{tab1}, we can observe that the BCL has a relatively limited effect in assisting models in most cases. This phenomenon is expected, as contrastive loss algorithms \cite{chopra2005learningr29,hoffer2015deepr30,ni2017finer32,sohn2016improvedr33,oh2016deepr34} typically have high requirements on the within-batch images and require careful selection of appropriate training images to ensure effectiveness. In contrast, to maintain the plug-and-play nature, we apply the BCL directly to randomly extracted mini-batches without imposing specific constraints on the selection of images. This practice inevitably limits performance improvement but is not the focus of our study. We will further analyze the complementary performance of GLS and BCL in Section \ref{GLS_BCL_experi}.

Meanwhile, we observe that BCL may lead to performance degradation in specific scenarios. For instance, when evaluating MobileNet on the MIT-67 dataset, all three BCL strategies, except for the "cosine\_mean\_inter" strategy, negatively affect the model's performance. However, it is important to note that the traditional CL resulted in an even greater performance degradation. This outcome could potentially be attributed to the inapplicability of the contrast loss strategy (batch level) to MobileNet's evaluation on the MIT-67 dataset. Nonetheless, in situations where such a conflict occurs, BCL applied under the "cosine\_mean\_inter" strategy manages to further improve the model performance, again demonstrating the significance of using the similarity prototype to provide prior knowledge during model training.

\subsection{Evaluation of Combining GLS and BCL}
\label{GLS_BCL_experi}
In this section, we employ both Gradient Label Softening (GLS) and Batch-level Contrastive Loss (BCL) strategies to assist model training. We evaluate their effectiveness on the MIT-67 and SUN397 datasets using multiple models, and the experimental results are presented in Table \ref{tab3}.

The result of Table \ref{tab3} shows that the combination of GLS and BCL significantly improves the accuracy of the trained models compared to the baseline. For example, on the MIT-67 dataset, utilizing both GLS and BCL strategies leads to a 4.552\% accuracy gain for MobileVIT, while on the SUN397 dataset, ConvNext achieves a 2.731\% accuracy improvement. Importantly, this improvement is achieved without any increase in network parameters or computational cost, which makes this result particularly remarkable. This outcome strongly underscores the importance and generalization capability of the proposed similarity prototype in improving model performance. 

In Table \ref{tab3}, we also explore the combination of LSR \cite{szegedy2016rethinkingr23} and CL \cite{chopra2005learningr29} for model training. It is observed that this combination offers only a marginal improvement in baseline accuracy compared to our GLS and BCL, and in some cases, it even negatively affects model performance. For example, when evaluating MobileNet on the MIT-67 dataset, combining GLS and BCL yields a 1.866\% increase in baseline performance. In contrast, combining LSR and CL results in a 0.388\% decrease. This contrast further confirms the significant role of our similarity prototype in providing prior knowledge during the model training process. Also, it supports our earlier observation (Section \ref{eval_BCL}) that MobileNet is less compatible with general metric learning strategies on the MIT-67 dataset.

\begin{table*}[!htbp]
    \centering
    \caption{Evaluation of combining BCL and GLS on MIT-67 and SUN397 datasets. The best result of each row is marked in bold.}
    \resizebox{1\textwidth}{!}{
    \begin{tabular}{cllccllll}
    \hline
        Dataset & Model & Param & Baseline & LSR + CL & \makecell[c]{GLS + BCL\textsubscript{Cosine} \\ mean\_inter} & \makecell[c]{GLS + BCL\textsubscript{Euclidean} \\ mean\_inter} & \makecell[c]{GLS + BCL\textsubscript{Cosine} \\ nonzero\_inter\_intra} & \makecell[c]{GLS + BCL\textsubscript{Euclidean} \\ nonzero\_inter\_intra} \\ \hline
        MIT-67 & ResNet50 & 25M & 76.493 & 76.791\textsubscript{\textcolor{red}{+0.328}} & \textbf{79.104}\textsubscript{\textcolor{red}{+2.611}} & 78.881 & 78.433 & 78.806  \\ 
        MIT-67 & VGG16 & 138M & 73.358  & 73.134\textsubscript{\textcolor{red}{-0.224}} & 75.0 & 75.522 & 73.881 & \textbf{75.746}\textsubscript{\textcolor{red}{+2.388}}  \\ 
        MIT-67 & MobileNet & 5.4M & 69.179 & 68.791\textsubscript{\textcolor{red}{-0.388}}  & \textbf{71.045}\textsubscript{\textcolor{red}{+1.866}} & 70.224 & 70.149 & 70.223  \\ 
        MIT-67 & ShuffleNet & 7M & 72.090 & 73.282\textsubscript{\textcolor{red}{+1.192}}  & \textbf{75.298}\textsubscript{\textcolor{red}{+3.208}} & 74.402 & 74.179 & 74.328  \\ 
        MIT-67 & MobileVit & 6M & 74.552 & 75.672\textsubscript{\textcolor{red}{+1.12}}  & \textbf{79.104}\textsubscript{\textcolor{red}{+4.552}} & 77.836 & 74.851 & 77.910  \\ 
        MIT-67 & SwinT & 87M & 88.681 & 88.507\textsubscript{\textcolor{red}{-0.174}}  & 89.03 & 89.104 & 89.03 & \textbf{89.179}\textsubscript{\textcolor{red}{+0.498}}  \\ 
        MIT-67 & VIT & 86M & 86.194 & 86.418\textsubscript{\textcolor{red}{+0.224}}  & \textbf{87.388}\textsubscript{\textcolor{red}{+1.194}} & 86.866 & 86.567 & 86.791  \\ 
        MIT-67 & ConvNext & 89M & 87.388 & 87.468\textsubscript{\textcolor{red}{+0.08}}  & \textbf{88.805}\textsubscript{\textcolor{red}{+1.417}} & 88.433 & 88.358 & 88.433  \\ \hline
        SUN397 & ResNet50 & 25M & 60.423 & 58.781\textsubscript{\textcolor{red}{-1.642}}  & 62.086 & 62.161 & \textbf{62.358}\textsubscript{\textcolor{red}{+1.935}} & 62.045  \\ 
        SUN397 & VGG16 & 138M & 57.204 & 57.471\textsubscript{\textcolor{red}{+0.267}}  & 58.519 & \textbf{58.982}\textsubscript{\textcolor{red}{+1.778}} & 57.516 & 58.872  \\ 
        SUN397 & MobileNet & 5.4M & 53.375 & 53.123\textsubscript{\textcolor{red}{-0.252}}  & 55.869 & 56.166 & 54.232 & \textbf{56.317}\textsubscript{\textcolor{red}{+2.942}}  \\ 
        SUN397 & ShuffleNet & 7M & 58.181 & 58.212\textsubscript{\textcolor{red}{+0.031}}  & 59.788 & 60.171 & 59.954 & \textbf{60.171}\textsubscript{\textcolor{red}{+1.99}}  \\ 
        SUN397 & MobileVit & 6M & 59.033 & 59.683\textsubscript{\textcolor{red}{+0.65}}  & 61.274 & 61.365 & 60.71 & \textbf{61.476}\textsubscript{\textcolor{red}{+2.433}}  \\ 
        SUN397 & SwinT & 87M & 75.063 & 76.151\textsubscript{\textcolor{red}{+1.088}}  & 77.118 & 77.083 & \textbf{77.294}\textsubscript{\textcolor{red}{+2.231}} & 77.108  \\ 
        SUN397 & VIT & 86M & 75.078 & 75.124\textsubscript{\textcolor{red}{+0.046}}  & 75.682 & 75.768 & \textbf{75.768}\textsubscript{\textcolor{red}{+0.69}} & 75.597  \\ 
        SUN397 & ConvNext & 89M & 74.675 & 76.156\textsubscript{\textcolor{red}{+1.481}}  & 77.259 & 77.259 & \textbf{77.406}\textsubscript{\textcolor{red}{+2.731}} & 77.229  \\ \hline
    \end{tabular}
    }
    \label{tab3}
\end{table*}

Next, we compare the best performance obtained using the GLS strategy, the best performance obtained using the BCL strategy, and the best performance obtained when the two strategies are applied simultaneously. These results are summarized in Table \ref{tab4}.

Table \ref{tab4} shows that the improvement in model performance using the BCL strategy, compared to the GLS strategy, is somewhat limited. This is because, to maintain the plug-and-play nature, we apply the BCL directly to randomly extracted mini-batches without imposing specific constraints on the selection of images. BCL differs from many contrastive loss algorithms \cite{chopra2005learningr29,hoffer2015deepr30,ni2017finer32,sohn2016improvedr33,oh2016deepr34}, which emphasizes the careful selection of appropriate training images for better performance results. Further analysis of Table \ref{tab4} reveals that, in most cases, there are complementary effects between GLS and BCL strategies. By incorporating these two training strategies simultaneously, we typically achieve higher accuracy than using either strategy alone. This finding further validates the potential of the proposed similarity prototype, indicating that there is still room for further development of the similarity prototype.

\begin{table}[!htbp]
    \centering
    \caption{Comparative ablation evaluation of three training strategies on MIT-67 and SUN397 datasets. The best result of each row is marked in bold.}
    \resizebox{0.6\textwidth}{!}{
    \begin{tabular}{cllcccc}
    \hline
        Dataset & Model & Param & Baseline & GLS & BCL & GLS + BCL \\ \hline
        MIT-67 & ResNet50 & 25M & 76.493 & \textbf{79.403} & 77.014 & 79.104  \\ 
        MIT-67 & VGG16 & 138M & 73.358 & 75.0 & 74.851 & \textbf{75.746}  \\ 
        MIT-67 & MobileNet & 5.4M & 69.179 & \textbf{71.119} & 69.478 & 71.045  \\ 
        MIT-67 & ShuffleNet & 7M & 72.090 & 74.925 & 74.104 & \textbf{75.298}  \\ 
        MIT-67 & MobileVit & 6M & 74.552 & 78.134 & 76.194 & \textbf{79.104}  \\ 
        MIT-67 & SwinT & 87M & 88.681 & 89.03 & \textbf{89.701} & 89.179  \\ 
        MIT-67 & VIT & 86M & 86.194 & 87.015 & 86.493 & \textbf{87.388}  \\ 
        MIT-67 & ConvNext & 89M & 87.388 & 88.657 & 88.657 & \textbf{88.805}  \\ \hline
        SUN397 & ResNet50 & 25M & 60.423 & 62.060 & 61.033 & \textbf{62.358}  \\ 
        SUN397 & VGG16 & 138M & 57.204 & 58.715 & 57.436 & \textbf{58.982}  \\ 
        SUN397 & MobileNet & 5.4M & 53.375 & 55.657 & 54.055 & \textbf{56.317}  \\ 
        SUN397 & ShuffleNet & 7M & 58.181 & 60.08 & 59.073 & \textbf{60.171}  \\ 
        SUN397 & MobileVit & 6M & 59.033 & 61.28 & 60.121 & \textbf{61.476}  \\ 
        SUN397 & SwinT & 87M & 75.063 & 77.194 & 75.95 & \textbf{77.294}  \\ 
        SUN397 & VIT & 86M & 75.078 & 75.758 & 75.521 & \textbf{75.768}  \\ 
        SUN397 & ConvNext & 89M & 74.675 & 77.274 & 75.592 & \textbf{77.406}  \\ \hline
    \end{tabular}
    }
    \label{tab4}
\end{table}

\subsection{Visualization on Places365-7 and Places365-14}
\begin{figure}[htbp]
    \centering
    \includegraphics[width=12cm]{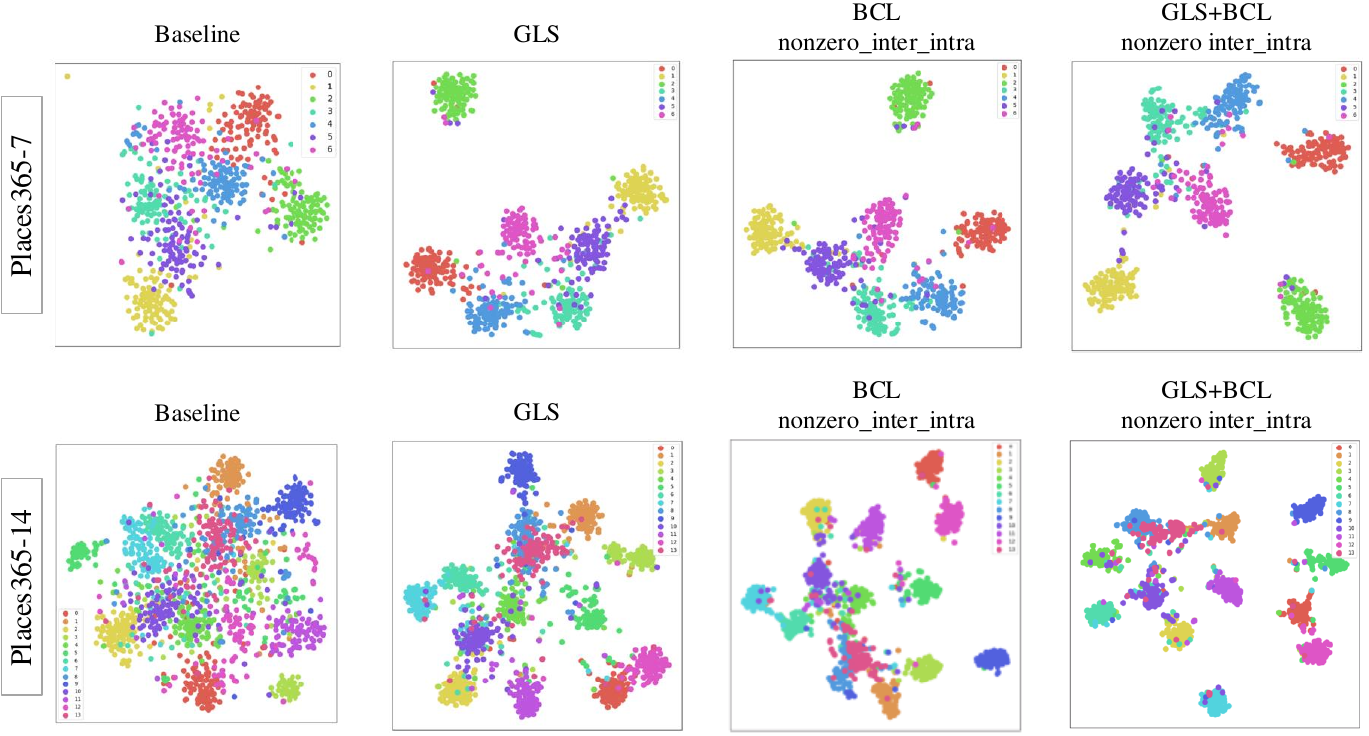}
    \caption{Visualization of the penultimate layer representation of ResNet50 on Places365-7 and Places365-14 validation set using t-SNE. Note that different colors represent different scene classes.}
    \label{fig8}
\end{figure}

To comprehensively comprehend the functioning of our method for model training, we select the visualizations produced by training ResNet50 with hard labels as a baseline. We then train it using the proposed strategy and evaluate its performance on two simplified versions of the validation set from the Places365 dataset, with the results shown in Table \ref{tab_places}. In addition, we extract the feature maps generated by the trained model. To provide a visual representation of these feature embeddings, we employ t-SNE, a dimensional reduction technique. By plotting the 2-dimensional feature representation of the embeddings in Fig. \ref{fig8}, we can effectively visualize the distribution of the images in the validation set. Each point on the plot corresponds to an individual image, and points sharing the same color indicate images belonging to the same category. Note that the proposed strategy used in this part is based on the cosine-based similarity prototype and the BCL strategy is "nonzero\_inter\_intra."

Upon observing Fig. \ref{fig8}, it can be concluded that employing any of our proposed strategies for model training has a significant positive impact on the network’s discriminative ability compared to the baseline. The trained model demonstrates enhanced capability in distinguishing different scene categories and exhibits superior aggregation of scene instances under the same category. Regarding the effect of feature visualization, we can see that both the proposed GLS and BCL strategies yield similar outcomes. Moreover, intuitively, the simultaneous application of these two auxiliary strategies results in a more discriminative network with clearer boundaries between different scene classes. This observation aligns with the experimental results presented in Table \ref{tab4} and \ref{tab_places}, providing additional evidence of the potential of the proposed similar prototype. In summary, based on the qualitative visualization and analysis presented in this section, it can be confidently affirmed that our proposed method exerts a highly impressive aiding effect on model training.

\begin{table}[htbp]
    \centering
    \caption{Evaluation on Places365-7 and Places365-14 datasets. The best result of each row is marked in bold.}
    \resizebox{0.6\textwidth}{!}{
    \begin{tabular}{llcccc}
    \hline
        Dataset & Model & Baseline & GLS & BCL & GLS+BCL \\ \hline
        Places365-7 & ResNet50 & 89 & 90.271 & 89.571 & \textbf{90.571} \\ 
        Places365-14 & ResNet50 & 85.786 & 86.163 & 86.357 & \textbf{86.571} \\ \hline
    \end{tabular}
    \label{tab_places}
    }
\end{table}

\subsection{Computational cost analysis}
To demonstrate the superior computational efficiency of our approach in real-world deployments, we conduct a comparative analysis to elucidate the difference in computational cost between local training and practical deployment. Specifically, we use ConvNext\_Base \cite{liu2022convnetr12} as the backbone network and VIT Adapter \cite{chen2023visionr41} as the semantic segmentation network, with the respective FLOPs and Inference throughput detailed in Table \ref{tab5}. The FLOPs values are sourced from the original papers. Regarding Inference throughput, considering that the two papers use different measurement devices, we run two modules on a 3090 GPU to measure Inference throughput for a fair comparison. Since GLS and BCL occupy very little FLOPs during training, we do not show them in Table \ref{tab5}. Note that since GLS and BCL are no longer implemented in the practical deployment, this omission does not affect our conclusions.

\begin{table}[htbp]
    \centering
    \caption{Comparison of computational cost during training and practical deployment}
    \resizebox{0.6\textwidth}{!}{ 
    \begin{tabular}{l|lc|lc}
    \hline
        Architecture & \makecell[c]{Training \\ FLOPs \\ (G)} & \makecell[c]{In practice \\ FLOPs \\ (G)} & \makecell[c]{Training \\ Throughout \\ (image / s)} & \makecell[c]{In practice \\ Throughout \\ (image / s)} \\ \hline
        \makecell[c]{Backbone \\ (ConvNext \cite{liu2022convnetr12})} & 15.4 & 15.4 & 115 & 115  \\ 
        Semantic Seg \cite{chen2023visionr41} & 473 & 0 & 0.09 & -  \\ 
        Total & 488.7 & \textbf{15.4} & $<$ 0.09 & \textbf{115} \\ \hline
    \end{tabular}
    }
    \label{tab5}
\end{table}

In Table \ref{tab5}, it can be observed that our approach relies only on the trained backbone network for real-time scene recognition. By effectively eliminating dependence on semantic segmentation, we achieve a significant decrease in the model's FLOPs, resulting in a substantial enhancement of inference throughput for practical applications. This result aligns with our initial expectations and strongly demonstrates the clear advantages of our approach when implemented in practice. 

Additionally, to the best of our knowledge, all existing object-assisted scene recognition methods require the object information extraction process during practical deployment. Given that the object information extraction process (Semantic Seg \cite{chen2023visionr41}) shown in Table \ref{tab5} has a throughput far lower than that of a typical image classification network (ConvNext \cite{liu2022convnetr12}), we can infer that our method will operate much faster in real-world deployments compared to other object-assisted methods. This further highlights the significant relevance of our proposed similarity prototype for object-assisted methods.

Furthermore, removing semantic segmentation in real-world applications alleviates the requirement about the Inference speed of the selected semantic segmentation technique. For instance, the VIT Adapter's limitations in inference speed make it unsuitable for real-world deployments. However, our method only uses this segmentation technique for model training, allowing us to leverage its precision advantages while maintaining practical applicability.

\subsection{Integration with existing semantic-guided method}

In this part, we apply the proposed similarity prototype-related strategies to an established semantic-guided scene recognition method, DGN-Net \cite{song2024interobject}. To ensure a fair comparison, all experiments in this part use training hyperparameters identical to those of the original DGN-Net. Initially, we reproduce the baseline performance of DGN-Net. Subsequently, we integrate our similarity prototype-related strategies into the training process of DGN-Net. The experimental results are displayed in Table \ref{tab_DGN}. Note that the proposed strategy used in this part is based on the Euclidean-based similarity prototype and the BCL strategy is "nonzero\_inter\_intra."

\begin{table}[htbp]
    \centering
    \caption{Integration evaluation in existing semantic-guided method}
    \resizebox{0.7\textwidth}{!}{ 
    \begin{tabular}{lllll}
    \hline
        Method & MIT-67 & SUN397 & Places365-7 & Places365-14 \\ \hline
        DGN-Net \cite{song2024interobject} & 90.373 & 79.765 & 94.286 & 89.914 \\ 
        DGN-Net* & 90.448 & 79.824 & 94 & 89.786 \\ 
        DGN-Net + GLS + BCL & \textbf{91.418} & \textbf{80.365} & \textbf{94.571} & \textbf{90.286} \\ \hline
        \multicolumn{5}{l}{\ {*} denotes the DGN-Net results as reproduced in our experiments.}
    \end{tabular}
    }
    \label{tab_DGN}
\end{table}

As shown in Table \ref{tab_DGN}, simply adding the proposed strategies on top of DGN-Net improves state-of-the-art performance across all evaluated datasets: MIT-67 (+0.97\%), SUN397 (+0.541\%), Places365-7 (+0.571\%) and Places365-14 (+0.5\%). These results further demonstrate the superiority and generalization of the prior knowledge provided by our similarity prototype for scene recognition. 

\subsection{Comparison with State-of-The-Art Methods}

We conduct a comparative analysis between our approach and existing state-of-the-art methods. As shown in Table \ref{tab6} and \ref{tab7}, these comparisons are conducted on several public datasets: MIT-67, SUN397, Places365-14, and Places365-7. We categorize these methods based on whether they require object information extraction techniques in practice (marked in the Semantic column). We use two models for comparison: one that does not require object information (ConcNext \cite{liu2022convnetr12} or ResNet50 \cite{he2016deepr5}) and another that does (DGN-Net \cite{song2024interobject}). By combining these models, we conduct a fair comparison between the proposed method and existing methods.

The results in Table \ref{tab6} and \ref{tab7} reveal that our method outperforms most previous methods. Specfically, methods \cite{xie2015hybridr13, xie2019hierarchicalr14, herranz2016scener15, song2017multir16,lin2022scener47} incorporating factors such as multi-scale information or multi-model combination for scene recognition. In contrast, our method, guided by inter-class correlation knowledge in the similarity prototype, achieves superior performance with only a single-branch and single-scale architecture. Notably, while CSDML \cite{WANG2022108589} also utilizes inter-class correlation to aid scene recognition, it adopts a traditional metric learning perspective to build class-level knowledge, ignoring the use of object information within scenes, and thus performs sub-optimally compared to our method. These comparisons not only prove the effectiveness of the proposed similarity prototype but also highlight the importance of incorporating object knowledge for scene recognition.

\begin{table}[htbp]
    \centering
    \caption{State-of-the-art results on MIT-67 and SUN397 dataset (\%).}
    \resizebox{0.6\textwidth}{!}{ 
    \begin{tabular}{lcccc}
    \hline
        Approaches & Input Size & Semantic & MIT-67 & SUN397 \\ \hline
        MLR+CFV+FCR \cite{xie2015hybridr13}   & 256 $\times$ 256 & -  & 82.24  & 64.53  \\ 
        NNSD+ICLC \cite{xie2019hierarchicalr14}   & 224 $\times$ 224 & - & 84.3  & 64.78  \\ 
        Multi-Scale CNN \cite{herranz2016scener15}   & 889 $\times$ 889 & - & 80.97  & 70.17  \\ 
        MP \cite{song2017multir16}   & 640 $\times$ 640 & -  & 86.9  & 72.6  \\ 
        SOSF+CFA+GAF \cite{sun2018fusingr17}   & 608 $\times$ 608  & \checkmark & 89.51  & 78.93  \\ 
        SDO \cite{cheng2018scener19}   & 224 $\times$ 224  & \checkmark & 86.76  & 73.41  \\ 
        SAS-Net \cite{lopez2020semanticr21}   & 224 $\times$ 224 & \checkmark  & 87.1  & 74.04  \\
        Li, et al. \cite{li2021place}  & 224 $\times$ 224 & \checkmark  & 91.26  & -  \\
        ARG-Net \cite{zeng2020amorphousr22}   & 448 $\times$ 448 & \checkmark  & 88.13  & 75.02  \\ 
        CSDML \cite{WANG2022108589}   & 224 $\times$ 224 & - & 88.28  & -  \\ 
        MR-Net \cite{lin2022scener47}   & 448 $\times$ 448  & - & 88.08  & 73.98  \\ 
        CSRRM \cite{song2023srrmr45}   & 224 $\times$ 224 & \checkmark & 88.731  & -  \\ 
        DGN-Net \cite{song2024interobject}  & 224 $\times$ 224 & \checkmark & \textcolor{red}{90.373}  & \textcolor{red}{79.765}  \\ \hline
        ConvNext + GLS + BCL  & 224 $\times$ 224 & - & \textbf{88.805}  & \textbf{77.406}  \\ 
        DGN-Net + GLS + BCL  & 224 $\times$ 224 & \checkmark & \textcolor{blue}{91.418}  & \textcolor{blue}{80.365}  \\ \hline
    \end{tabular}}
    \label{tab6}
\end{table}

\begin{table}[htbp]
    \footnotesize
    \centering
    \caption{State-of-the-art results on Places365-14 and Places365-7 dataset (\%).}
    \begin{adjustbox}{width=0.7\textwidth}
    \begin{tabular}{lcccc}
    \hline
        Approaches  & Input Size & Semantic & Places365-14 & Places365-7 \\ \hline
        Word2Vec \citep{r_Word2Vec}  & 224 $\times$ 224 & \checkmark & 83.7 & -  \\ 
        Deduce \citep{r_Deduce} & 224 $\times$ 224 & \checkmark & - & 88.1  \\ 
        BORM-Net \citep{r_BORM} & 224 $\times$ 224 & \checkmark & 85.8 & 90.1  \\ 
        OTS-Net \citep{r_OTS} & 224 $\times$ 224 & \checkmark & 85.9 & 90.1  \\
        CSRRM \citep{song2023srrmr45} & 224 $\times$ 224 & \checkmark & 88.714 & 93.429  \\ 
        DGN-Net \cite{song2024interobject}  & 224 $\times$ 224 & \checkmark & \textcolor{red}{89.914}  & \textcolor{red}{94.286}  \\ \hline
        ResNet50 + GLS + BCL  & 224 $\times$ 224 & - & \textbf{86.571}  & \textbf{90.571}  \\ 
        DGN-Net + GLS + BCL & 224 $\times$ 224 & \checkmark & \textcolor{blue}{90.286} & \textcolor{blue}{94.571}  \\ \hline
    \end{tabular}
    \end{adjustbox}
    \label{tab7}
\end{table}

The focus then shifts to object information-assisted methods. Methods \cite{cheng2018scener19, r_Word2Vec, r_BORM} also use statistics to process object information within scenes to aid scene recognition. However, they rely on inter-object correlation to enhance feature discriminability, making them dependent on the precision of object information extraction techniques. In particular, their performance is more restricted when the precision of object information extraction is limited for a specific image. In contrast, our approach begins from a more generalized perspective, i.e., using statistics combined with object information to infer inter-class correlations. This knowledge can fundamentally improve the model's discriminability during training, making it no longer limited by the precision of object information extraction techniques in practice, and thus achieving superior performance.

Overall, all previous object-assisted methods \cite{cheng2018scener19,li2021place,lopez2020semanticr21,zeng2020amorphousr22,song2023srrmr45, song2024interobject, r_Word2Vec, r_Deduce, r_BORM, r_OTS} require substantial computational costs to extract object information within scenes in practical applications. In contrast, our method uses object knowledge only during training to improve the network's feature extraction ability, avoiding the computational burden of object information extraction in practical applications. Nevertheless, despite no longer using object knowledge in practice, our method still achieves superior recognition performance than most object-assisted methods. This finding once again emphasizes the superiority and practicality of our similarity prototype.

Furthermore, since our approach relies solely on backbone networks for scene recognition, some semantic-guided methods, such as DGN-Net \cite{song2024interobject}, inevitably achieve higher accuracy. However, incorporating our proposed strategies into DGN-Net can further improve its performance across all evaluated datasets. This result again demonstrates the potential of our similarity prototype, which can serve as an auxiliary strategy to further enhance state-of-the-art methods.

In summary, Table \ref{tab6} and \ref{tab7} provide significant evidence for the superiority and practicality of the proposed similarity prototype in advancing scene recognition.
 
\section{Conclusions}
\label{Conclusions}
In this paper, we propose embedding semantic knowledge into a similarity prototype, which can be used to supervise model training without adding any network parameter. Two strategies are proposed to fully unleash the potential of our similarity prototype for assisting scene recognition: Gradient Label Softening and Batch-level Contrastive Loss. The former embeds our prototype in softened labels and uses a confidence gradient strategy to guide network training. The latter uses our prototype to measure the similarity requirements of the inter-class and intra-class samples in each mini-batch. Positive results are achieved by employing both of the two strategies to assist scene recognition. We evaluate our approach through comprehensive experiments on three widely recognized datasets: MIT-67, SUN397, and Paces365. The results indicate that the proposed similarity prototype effectively enhances the network performance, all while avoiding any additional network parameters or computational costs in practical deployment.

Several interesting directions can be followed up, which are not covered by this paper. For instance, one promising direction could involve exploring the utilization of similarity prototypes for enhancing scene recognition through self-knowledge distillation. In addition, the BCL strategy designed in this work is relatively simple and has limited performance gains on some models (e.g., MobileNet in Section \ref{eval_BCL}). It is worth exploring and designing contrastive loss strategies that are more suitable for the proposed similarity prototype, which may lead to more significant performance gains.

Furthermore, experimental evidence shows that introducing our similarity prior knowledge markedly enhances the performance over two baseline strategies, Label Smoothing Regularization (LSR) and Contrastive Loss (CL). Intuitively, integrating our similarity prototype into more sophisticated metric learning algorithms is likely to yield even better results, meriting further exploration.

Also, this study uses all semantic objects from the ADE20K dataset to generate the similarity prototype. However, objects with low discriminative significance may contribute little. Therefore, carefully selecting the object categories used to generate the similarity prototype to eliminate redundancy could potentially enhance its performance for scene recognition.

As the first approach successfully leveraging object information for enhancing scene recognition performance without imposing additional computational burdens, we anticipate that this work could inspire future research to develop more efficient object-assisted scene recognition methods.

\section*{Acknowledgment}
This work was jointly supported by the Key Development Program for Basic Research of Shandong Province under Grant ZR2019ZD07, the National Natural Science Foundation of China-Regional Innovation Development Joint Fund Project under Grant U21A20486, the Fundamental Research Funds for the Central Universities under Grant 2022JC011, the Natural Science Youth Foundation of Shandong Province under Grant ZR2023QF055.



\bibliographystyle{elsarticle-num} 
\bibliography{refs.bib}

\begin{thebibliography}{10}
\expandafter\ifx\csname url\endcsname\relax
  \def\url#1{\texttt{#1}}\fi
\expandafter\ifx\csname urlprefix\endcsname\relax\def\urlprefix{URL }\fi
\expandafter\ifx\csname href\endcsname\relax
  \def\href#1#2{#2} \def\path#1{#1}\fi

\bibitem{xie2020scene}
L.~Xie, F.~Lee, L.~Liu, K.~Kotani, Q.~Chen, Scene recognition: A comprehensive survey, Pattern Recognition 102 (2020) 107205.

\bibitem{he2016deepr5}
K.~He, X.~Zhang, S.~Ren, J.~Sun, Deep residual learning for image recognition, in: Proceedings of the IEEE conference on computer vision and pattern recognition, 2016, pp. 770--778.

\bibitem{howard2019searchingr7}
A.~Howard, M.~Sandler, G.~Chu, L.-C. Chen, B.~Chen, M.~Tan, W.~Wang, Y.~Zhu, R.~Pang, V.~Vasudevan, et~al., Searching for mobilenetv3, in: Proceedings of the IEEE/CVF international conference on computer vision, 2019, pp. 1314--1324.

\bibitem{liu2021swinr10}
Z.~Liu, Y.~Lin, Y.~Cao, H.~Hu, Y.~Wei, Z.~Zhang, S.~Lin, B.~Guo, Swin transformer: Hierarchical vision transformer using shifted windows, in: Proceedings of the IEEE/CVF international conference on computer vision, 2021, pp. 10012--10022.

\bibitem{liu2022convnetr12}
Z.~Liu, H.~Mao, C.-Y. Wu, C.~Feichtenhofer, T.~Darrell, S.~Xie, A convnet for the 2020s, in: Proceedings of the IEEE/CVF conference on computer vision and pattern recognition, 2022, pp. 11976--11986.

\bibitem{xie2015hybridr13}
G.-S. Xie, X.-Y. Zhang, S.~Yan, C.-L. Liu, Hybrid cnn and dictionary-based models for scene recognition and domain adaptation, IEEE Transactions on Circuits and Systems for Video Technology 27~(6) (2015) 1263--1274.

\bibitem{cheng2018scener19}
X.~Cheng, J.~Lu, J.~Feng, B.~Yuan, J.~Zhou, Scene recognition with objectness, Pattern Recognition 74 (2018) 474--487.

\bibitem{sun2018fusingr17}
N.~Sun, W.~Li, J.~Liu, G.~Han, C.~Wu, Fusing object semantics and deep appearance features for scene recognition, IEEE Transactions on Circuits and Systems for Video Technology 29~(6) (2018) 1715--1728.

\bibitem{song2019imager20}
X.~Song, S.~Jiang, B.~Wang, C.~Chen, G.~Chen, Image representations with spatial object-to-object relations for rgb-d scene recognition, IEEE Transactions on Image Processing 29 (2019) 525--537.

\bibitem{li2021place}
P.~Li, X.~Li, X.~Li, H.~Pan, M.~O. Khyam, M.~Noor-A-Rahim, S.~S. Ge, Place perception from the fusion of different image representation, Pattern Recognition 110 (2021) 107680.

\bibitem{lopez2020semanticr21}
A.~L{\'o}pez-Cifuentes, M.~Escudero-Vinolo, J.~Besc{\'o}s, {\'A}.~Garc{\'\i}a-Mart{\'\i}n, Semantic-aware scene recognition, Pattern Recognition 102 (2020) 107256.

\bibitem{zeng2020amorphousr22}
H.~Zeng, X.~Song, G.~Chen, S.~Jiang, Amorphous region context modeling for scene recognition, IEEE Transactions on Multimedia 24 (2022) 141--151.

\bibitem{song2023srrmr45}
C.~Song, X.~Ma, Srrm: Semantic region relation model for indoor scene recognition, in: 2023 International Joint Conference on Neural Networks (IJCNN), 2023, pp. 01--08.

\bibitem{hou2024network}
Y.~Hou, Z.~Ma, C.~Liu, Z.~Wang, C.~C. Loy, Network pruning via resource reallocation, Pattern Recognition 145 (2024) 109886.

\bibitem{qiu2021essence}
J.~Qiu, Y.~Yang, X.~Wang, D.~Tao, Scene essence, in: 2021 IEEE/CVF Conference on Computer Vision and Pattern Recognition (CVPR), 2021, pp. 8318--8329.
\newblock \href {https://doi.org/10.1109/CVPR46437.2021.00822} {\path{doi:10.1109/CVPR46437.2021.00822}}.

\bibitem{szegedy2016rethinkingr23}
C.~Szegedy, V.~Vanhoucke, S.~Ioffe, J.~Shlens, Z.~Wojna, Rethinking the inception architecture for computer vision, in: Proceedings of the IEEE conference on computer vision and pattern recognition, 2016, pp. 2818--2826.

\bibitem{chopra2005learningr29}
S.~Chopra, R.~Hadsell, Y.~LeCun, Learning a similarity metric discriminatively, with application to face verification, in: 2005 IEEE computer society conference on computer vision and pattern recognition (CVPR'05), Vol.~1, IEEE, 2005, pp. 539--546.

\bibitem{quattoni2009recognizingr38}
A.~Quattoni, A.~Torralba, Recognizing indoor scenes, in: 2009 IEEE conference on computer vision and pattern recognition, IEEE, 2009, pp. 413--420.

\bibitem{xiao2010sunr39}
J.~Xiao, J.~Hays, K.~A. Ehinger, A.~Oliva, A.~Torralba, Sun database: Large-scale scene recognition from abbey to zoo, in: 2010 IEEE computer society conference on computer vision and pattern recognition, IEEE, 2010, pp. 3485--3492.

\bibitem{zhou2017placesr40}
B.~Zhou, A.~Lapedriza, A.~Khosla, A.~Oliva, A.~Torralba, Places: A 10 million image database for scene recognition, IEEE transactions on pattern analysis and machine intelligence 40~(6) (2017) 1452--1464.

\bibitem{simonyan2014veryr6}
K.~Simonyan, A.~Zisserman, Very deep convolutional networks for large-scale image recognition, arXiv preprint arXiv:1409.1556 (2014).

\bibitem{ma2018shufflenetr8}
N.~Ma, X.~Zhang, H.-T. Zheng, J.~Sun, Shufflenet v2: Practical guidelines for efficient cnn architecture design, in: Proceedings of the European conference on computer vision (ECCV), 2018, pp. 116--131.

\bibitem{mehta2021mobilevitr9}
S.~Mehta, M.~Rastegari, Mobilevit: light-weight, general-purpose, and mobile-friendly vision transformer, in: ICLR, 2022.

\bibitem{dosovitskiy2020imager11}
A.~Dosovitskiy, L.~Beyer, A.~Kolesnikov, D.~Weissenborn, X.~Zhai, T.~Unterthiner, M.~Dehghani, M.~Minderer, G.~Heigold, S.~Gelly, et~al., An image is worth 16x16 words: Transformers for image recognition at scale, in: ICLR, 2021.

\bibitem{xie2019hierarchicalr14}
L.~Xie, F.~Lee, L.~Liu, Z.~Yin, Q.~Chen, Hierarchical coding of convolutional features for scene recognition, IEEE Transactions on Multimedia 22~(5) (2019) 1182--1192.

\bibitem{herranz2016scener15}
L.~Herranz, S.~Jiang, X.~Li, Scene recognition with cnns: objects, scales and dataset bias, in: Proceedings of the IEEE Conference on Computer Vision and Pattern Recognition, 2016, pp. 571--579.

\bibitem{song2017multir16}
X.~Song, S.~Jiang, L.~Herranz, Multi-scale multi-feature context modeling for scene recognition in the semantic manifold, IEEE Transactions on Image Processing 26~(6) (2017) 2721--2735.

\bibitem{lin2022scener47}
C.~Lin, F.~Lee, L.~Xie, J.~Cai, H.~Chen, L.~Liu, Q.~Chen, Scene recognition using multiple representation network, Applied Soft Computing 118 (2022) 108530.

\bibitem{muller2019doesr25}
R.~M{\"u}ller, S.~Kornblith, G.~E. Hinton, When does label smoothing help?, Advances in neural information processing systems 32 (2019).

\bibitem{reed2014trainingr24}
S.~Reed, H.~Lee, D.~Anguelov, C.~Szegedy, D.~Erhan, A.~Rabinovich, Training deep neural networks on noisy labels with bootstrapping, arXiv preprint arXiv:1412.6596 (2014).

\bibitem{li2019reconstructionr26}
C.~Li, C.~Liu, L.~Duan, P.~Gao, K.~Zheng, Reconstruction regularized deep metric learning for multi-label image classification, IEEE transactions on neural networks and learning systems 31~(7) (2019) 2294--2303.

\bibitem{zhang2021delvingr27}
C.-B. Zhang, P.-T. Jiang, Q.~Hou, Y.~Wei, Q.~Han, Z.~Li, M.-M. Cheng, Delving deep into label smoothing, IEEE Transactions on Image Processing 30 (2021) 5984--5996.

\bibitem{gao2022labelr28}
F.~Gao, X.~Luo, Z.~Yang, Q.~Zhang, Label smoothing and task-adaptive loss function based on prototype network for few-shot learning, Neural Networks 156 (2022) 39--48.

\bibitem{kaya2019deepr31}
M.~Kaya, H.~{\c{S}}. Bilge, Deep metric learning: A survey, Symmetry 11~(9) (2019) 1066.

\bibitem{hoffer2015deepr30}
E.~Hoffer, N.~Ailon, Deep metric learning using triplet network, in: Similarity-Based Pattern Recognition: Third International Workshop, SIMBAD 2015, Copenhagen, Denmark, October 12-14, 2015. Proceedings 3, Springer, 2015, pp. 84--92.

\bibitem{ni2017finer32}
J.~Ni, J.~Liu, C.~Zhang, D.~Ye, Z.~Ma, Fine-grained patient similarity measuring using deep metric learning, in: Proceedings of the 2017 ACM on Conference on Information and Knowledge Management, 2017, pp. 1189--1198.

\bibitem{sohn2016improvedr33}
K.~Sohn, Improved deep metric learning with multi-class n-pair loss objective, Advances in neural information processing systems 29 (2016).

\bibitem{oh2016deepr34}
H.~Oh~Song, Y.~Xiang, S.~Jegelka, S.~Savarese, Deep metric learning via lifted structured feature embedding, in: Proceedings of the IEEE conference on computer vision and pattern recognition, 2016, pp. 4004--4012.

\bibitem{gonzalez2022guidedr36}
J.~Gonzalez-Zapata, I.~Reyes-Amezcua, D.~Flores-Araiza, M.~Mendez-Ruiz, G.~Ochoa-Ruiz, A.~Mendez-Vazquez, Guided deep metric learning, in: Proceedings of the IEEE/CVF Conference on Computer Vision and Pattern Recognition, 2022, pp. 1481--1489.

\bibitem{zhang2023graphr37}
C.-Y. Zhang, H.-C. Cai, C.~P. Chen, Y.-N. Lin, W.-P. Fang, Graph representation learning with adaptive metric, IEEE Transactions on Network Science and Engineering (2023).

\bibitem{chen2023visionr41}
Z.~Chen, Y.~Duan, W.~Wang, J.~He, T.~Lu, J.~Dai, Y.~Qiao, Vision transformer adapter for dense predictions, in: ICLR, 2023.

\bibitem{zhou2017scener42}
B.~Zhou, H.~Zhao, X.~Puig, S.~Fidler, A.~Barriuso, A.~Torralba, Scene parsing through ade20k dataset, in: Proceedings of the IEEE conference on computer vision and pattern recognition, 2017, pp. 633--641.

\bibitem{russakovsky2015imagenetr43}
O.~Russakovsky, J.~Deng, H.~Su, J.~Krause, S.~Satheesh, S.~Ma, Z.~Huang, A.~Karpathy, A.~Khosla, M.~Bernstein, et~al., Imagenet large scale visual recognition challenge, International journal of computer vision 115 (2015) 211--252.

\bibitem{kingma2014adamr44}
D.~P. Kingma, J.~Ba, Adam: A method for stochastic optimization, in: ICLR, 2015.

\bibitem{song2024interobject}
C.~Song, H.~Wu, X.~Ma, Inter-object discriminative graph modeling for indoor scene recognition (2024).
\newblock \href {http://arxiv.org/abs/2311.05919} {\path{arXiv:2311.05919}}.

\bibitem{WANG2022108589}
C.~Wang, G.~Peng, B.~{De Baets}, Class-specific discriminative metric learning for scene recognition, Pattern Recognition 126 (2022) 108589.

\bibitem{r_Word2Vec}
B.~X. Chen, R.~Sahdev, D.~Wu, X.~Zhao, M.~Papagelis, J.~K. Tsotsos, Scene classification in indoor environments for robots using context based word embeddings, in: 2018 International Conference on Robotics and Automation (ICRA) Workshop, 2018.

\bibitem{r_Deduce}
A.~Pal, C.~Nieto-Granda, H.~I. Christensen, Deduce: Diverse scene detection methods in unseen challenging environments, in: 2019 IEEE/RSJ International Conference on Intelligent Robots and Systems (IROS), IEEE, 2019, pp. 4198--4204.

\bibitem{r_BORM}
L.~Zhou, J.~Cen, X.~Wang, Z.~Sun, T.~L. Lam, Y.~Xu, Borm: Bayesian object relation model for indoor scene recognition, in: 2021 IEEE/RSJ International Conference on Intelligent Robots and Systems (IROS), IEEE, 2021, pp. 39--46.

\bibitem{r_OTS}
B.~Miao, L.~Zhou, A.~S. Mian, T.~L. Lam, Y.~Xu, Object-to-scene: Learning to transfer object knowledge to indoor scene recognition, in: 2021 IEEE/RSJ International Conference on Intelligent Robots and Systems (IROS), IEEE, 2021, pp. 2069--2075.

\end{thebibliography}





\end{document}